\documentclass{article} 
\usepackage{iclr2021_conference,times}

\usepackage{times}
\usepackage{latexsym}

\usepackage[colorinlistoftodos]{todonotes}
\usepackage{microtype}
\usepackage{amsmath}
\usepackage{amssymb}
\usepackage{ctable}
\usepackage{enumitem}
\usepackage{pifont}
\usepackage{amsfonts}
\usepackage{makecell}
\usepackage{url}
\usepackage{graphicx}
\usepackage{footnote}
\usepackage{soul}
\usepackage{tablefootnote}
\usepackage{xspace}

\usepackage{subfig}
\usepackage{adjustbox}

\usepackage{hyperref}

\makesavenoteenv{tabular}
\makesavenoteenv{table}
\graphicspath{{images/}}

\iclrfinalcopy

\newcommand{\wikisql}{\textsc{WikiSQL}}
\newcommand{\spider}{\textsc{Spider}}
\newcommand{\wikitable}{\textsc{WikiTableQuestions}}

\newcommand{\tablesp}{table semantic parsing\xspace}

\newcommand{\roberta}{RoBERTa}
\newcommand{\bert}{\textsc{BERT}}

\newcommand{\tablebert}{\textsc{Grappa}}

\newcommand{\vic}[1]{\textbf{\textcolor{magenta}{VL: #1}}}
\newcommand{\drr}[1]{\textbf{\textcolor{purple}{DR: #1}}}
\newcommand{\ty}[1]{\textbf{\textcolor{blue}{TY: #1}}}

\newcommand{\hide}[1]{}

\title{\tablebert: Grammar-Augmented Pre-Training for Table Semantic Parsing}

\author{

\makecell{Tao Yu$^\dag$\thanks{~~This work was mostly done during Tao and Bailin's internship at Salesforce Research. Victoria is now at Facebook AI.}\ \ , Chien-Sheng Wu$^\S$, Xi Victoria Lin$^\S{^\ast}$, Bailin Wang$^\ddag{^\ast}$, Yi Chern Tan$^\dag$, 
Xinyi Yang$^\S$, \\ Dragomir Radev$^\dag{^\S}$, Richard Socher$^\S$, Caiming Xiong$^\S$} \\

\centerline{ 
$^\S$ Salesforce Research, 
$^\dag$ Yale University, $^\ddag$University of Edinburgh 
}\\

\centerline{ 
\tt{\{tao.yu, yichern.tan,dragomir.radev\}@yale.edu},
\tt{bailin.wang@ed.ac.uk}} \\

\centerline{ 
\tt{\{wu.jason,x.yang,rsocher,cxiong\}@salesforce.com},
\tt{victorialin@fb.com}}
}

\date{}

\begin{document}
\maketitle

\begin{abstract}
We present \tablebert{}, an effective pre-training approach for \tablesp that learns a compositional inductive bias in the joint representations of textual and tabular data.
We construct synthetic question-SQL pairs over high-quality tables via a synchronous context-free grammar (SCFG).
We pre-train \tablebert{} on the synthetic data to inject important structural properties commonly found in table semantic parsing into the pre-training language model.
To maintain the model's ability to represent real-world data, we also include 
masked language modeling (MLM) on several existing table-and-language datasets 
to regularize our pre-training process. 
Our proposed pre-training strategy is much data-efficient.
When incorporated with strong base semantic parsers, \tablebert{} achieves new state-of-the-art results on four popular fully supervised and weakly supervised table semantic parsing tasks.
The pre-trained embeddings can be downloaded at \url{https://huggingface.co/Salesforce/grappa_large_jnt}. 

\end{abstract}
\section{Introduction}
\label{introduction}


Tabular data serve as important information source for human decision makers in many domains, such as finance, health care, retail and so on. While tabular data can be efficiently accessed via the structured query language (SQL), 
a natural language interface 
allows such data to be more accessible for a wider range of non-technical users. As a result, table semantic parsing that maps natural language queries over tabular data to formal programs has drawn significant attention in recent years. 


Recent  pre-trained language models (LMs) such as BERT~\citep{Devlin2019BERTPO} and RoBERTa~\citep{Liu2019RoBERTaAR} achieve tremendous success on a spectrum of natural language processing tasks, including semantic parsing~\citep{Zettlemoyer05,Zhong2017,Yu18emnlp}.
These advances have shifted the focus from building domain-specific semantic parsers~\citep{Zettlemoyer05,artzi13,Berant14,li2014constructing} to cross-domain semantic parsing~\citep{Zhong2017,Yu18emnlp,Herzig2018,dong18,wang-etal-2020-rat,LinRX2020:BRIDGE}.

Despite such significant gains, the overall performance on complex benchmarks such \spider{}~\citep{Yu18emnlp} and \wikitable{ benchmarks} are still limited, even when integrating representations of current pre-trained
language models. 
As such tasks requires generalization to new databases/tables and more complex programs (e.g., SQL), 
we hypothesize that current pre-trained language models are not sufficient for such tasks. 
First, language models pre-trained using unstructured text data such as Wikipedia and Book Corpus 
are exposed to a significant domain shift when directly applied to table semantic parsing, where jointly modeling the relation between utterances and structural tables is crucial.
Second, conventional pre-training objectives does not consider the underlying compositionality of data (e.g., questions and SQLs) from table semantic parsing.
To close this gap, we seek to learn contextual representations jointly from structured tabular data and unstructured natural language sentences, with objectives oriented towards table semantic parsing. 

In this paper, we propose a novel grammar-augmented pre-training framework for table semantic parsing (\tablebert{}).
Inspired by previous work on data synthesis for semantic parsing~\citep{Berant14,wang2015building,jia2016,Herzig2018,andreas-2020-good}, we induce a 
synchronous context-free grammar (SCFG) specific to mapping natural language to SQL queries from existing text-to-SQL datasets, 
which covers 
most commonly used question-SQL patterns.
As shown in Figure~\ref{fig:intro_example}, from a text-to-SQL example 
we can create a question-SQL template by abstracting over mentions of schema components (tables and fields), values, and SQL operations.
By executing this template on randomly selected tables we can create a large number of synthetic question-SQL pairs.
We train \tablebert{} on these synthetic question-SQL pairs and their corresponding tables using a novel text-schema linking objective that predicts the syntactic role of a table column in the SQL for each pair. This way we encourage the model to identify table schema components that can be grounded to logical form constituents, which is critical for most table semantic parsing tasks.

\begin{figure}[t!]
    \centering
    \includegraphics[width=0.9 \textwidth]{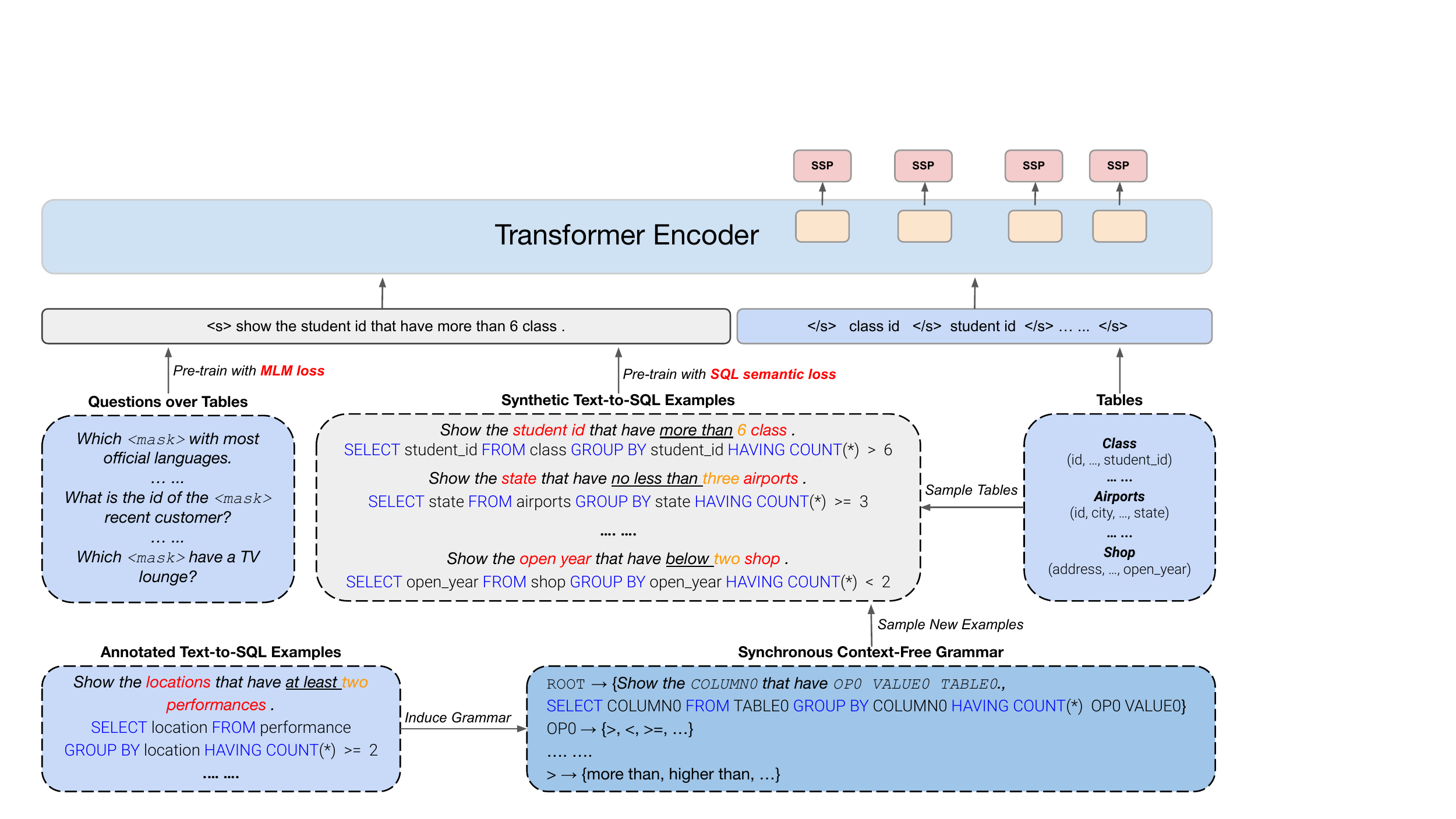}
    \caption{An overview of \tablebert{} pre-training approach. We first induce a SCFG given some examples in \spider{}. We then sample from this grammar given a large amount of tables to generate new synthetic examples. Finally, \tablebert{} is pre-trained on the synthetic data using SQL semantic loss and a small amount of table related utterances using MLM loss.}
\label{fig:intro_example}
\vspace{-2mm}
\end{figure}

To prevent overfitting to the synthetic data,
we include the masked-language modelling (MLM) loss on 
several large-scale, high-quality table-and-language datasets 
and carefully balance between preserving the original natural language representations and injecting the compositional inductive bias through our synthetic data.
We pre-train \tablebert{} using 475k synthetic examples and 391.5k examples from existing table-and-language datasets. 
Our approach dramatically reduces the training time and GPU cost.
We evaluate on four popular semantic parsing benchmarks in both fully supervised and weakly supervised settings. 
\tablebert{} consistently achieves new state-of-the-art results on all of them, significantly outperforming all previously reported results.




\hide{
The goal of table-based semantic parsing is to map natural language utterances into machine interpretable executable programs such as SQL queries.
Most previous work \citep{zelle96,Zettlemoyer05,artzi13,Berant14,li2014constructing} has focused on building a semantic parser that works for one single domain or different parsers for different domains, but they are unable to generalize to an unseen domain. \ty{modify and delete}
Such systems have become impractical as the number of domains grows (change). 

\begin{figure}[t!]
    \centering
    \includegraphics[width=0.9 \textwidth]{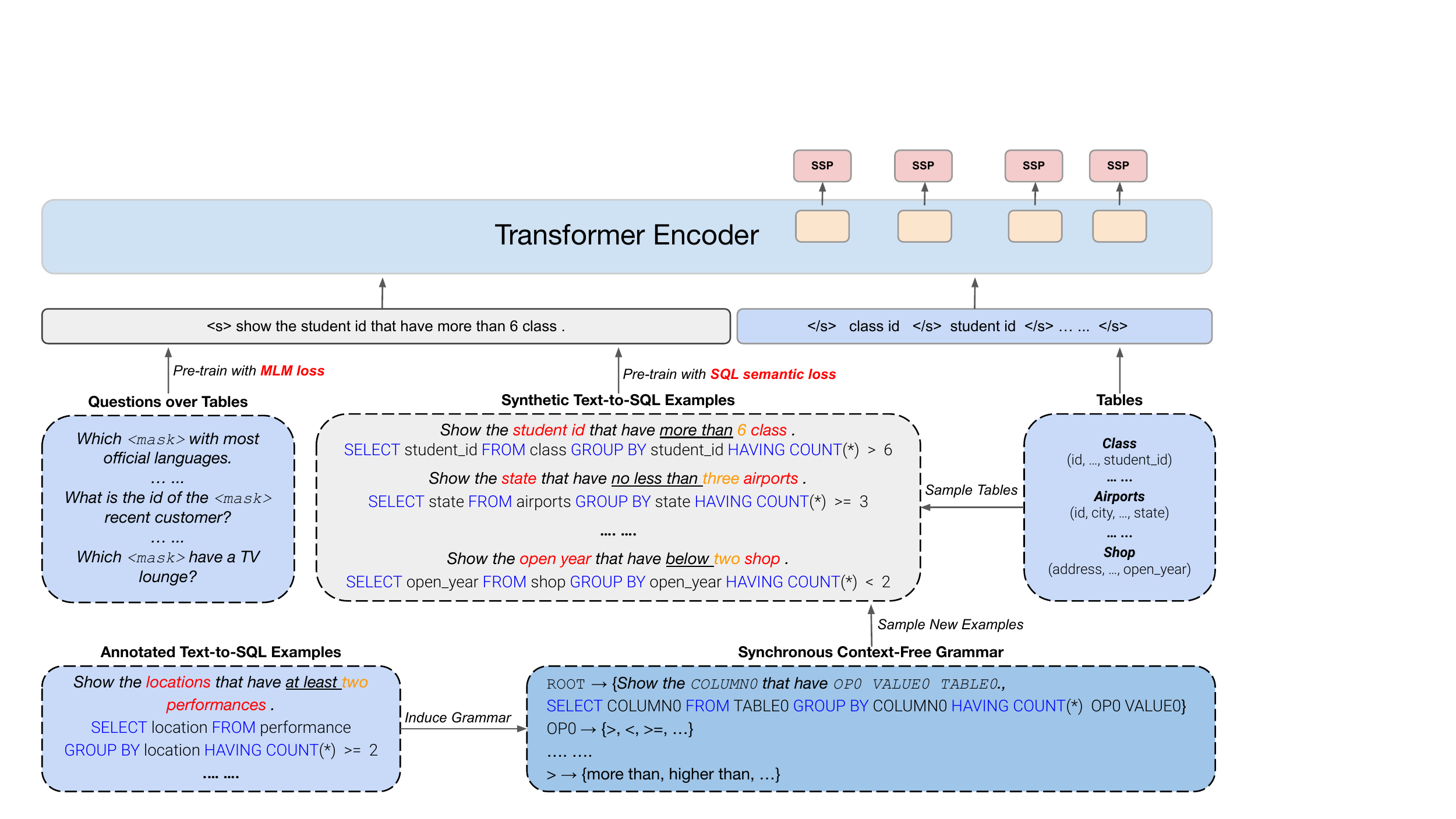}
    \caption{An overview of \tablebert{} pre-training approach. We first induce a SCFG given some examples in \spider{}. We then sample from this grammar given a large amount of tables to generate new synthetic examples. Finally, \tablebert{} is pre-trained on the synthetic data using SQL semantic loss and a small amount of table related utterances using MLM loss.}
\label{fig:intro_example}
\vspace{-2mm}
\end{figure}


Recent work that adopts deep neural models~\citep{Zhong2017,Yu18emnlp,Herzig2018} has shifted the focus to cross-domain semantic parsing.
Models using BERT~\citep{Devlin2019BERTPO}/RoBERTa~\citep{Liu2019RoBERTaAR}
perform significantly better than non-BERT approaches on different text-to-SQL benchmarks such as \wikisql{} and \spider{}.
Notably on \wikisql{}, \citet{Hwang2019ACE} are able to match the upper-bound performance simply by incorporating BERT as the encoder. 
However, there is still much space for improvement on the \spider{} task, which contains databases with multiple tables and SQL-complex questions.
One possible reason is that those pre-trained models are learned using unstructured text data such as Wikipedia and Book Corpus, which may still have a significant domain shift while applying to table-related tasks.
\ty{no need to mention this}


In this paper, we aim to have a pre-trained language model for table parsing that is jointly learned natural language sentence and table context.
To obtain high quality data, we first aggregate six table-related datasets, including table question answering, table-to-text generation, table fact checking, and text-to-SQL tasks.
In total, there are around 383K training pairs involving over 100K tables.
Moreover, we propose a based on \ty{to modify}

By incorporating BERT as the encoder, we are able to see vast improvements in (1) and (2).
However, because BERT is trained on free text, we do not see similar improvements in (3), where mentions of columns and values in the question can be noisy, ambiguous, and domain-specific.
On \wikisql{}, which has over 80k examples with simple SQL logic patterns ({\small \tt SELECT} and {\small \tt WHERE}), BERT is able to successfully capture the logic mapping via pre-training.
However, we do not have access to such large-scale data with more complex SQL logic patterns. 
For example, \spider{}, the largest dataset containing complex question-query pairs, only includes fewer than 1k examples for each SQL pattern with complex logic.\vic{What does ``fewer than 1k examples for each SQL pattern'' mean?} \ty{shorten this part}

\begin{table*}[ht!]
\centering
\scalebox{0.80}{
\begin{tabular}{ccccc}
\Xhline{2\arrayrulewidth}
Task \& Dataset & \# Examples & Resource & Annotation & Cross-domain\\\hline
\textsc{Spider} \cite{Yu18emnlp} & 10,181  & database & SQL & \checkmark\\
Fully-sup. \textsc{WikiSQL} \cite{Zhong2017} & 80,654  & single table & SQL & \checkmark\\\hline
\textsc{WikiTableQuestions} \cite{pasupat2015compositional} & 2,2033 & single table & answer & \checkmark \\
Weakly-sup. \textsc{WikiSQL} \cite{Zhong2017} & 80,654  & single table & answer & \checkmark\\\hline
\Xhline{2\arrayrulewidth}
\end{tabular}}
\caption{Overview of four table-based semantic parsing and question answering datasets in fully-supervised (top) and weakly-supervised (bottom) setting used in this paper. More details in Section \ref{exp:experiment}}
\label{tb:data_stats}
\vspace{-3mm}
\end{table*}

To this end, inspired by previous work~\citep{Berant14,wang2015building,jia2016,Yu18syntax} who use context-free grammars to automatically augment the data for semantic parsing, we develop a SQL-specific synchronous context-free grammar (SCFG) which covers a set of the most frequently used SQL patterns and their corresponding questions in \spider{}.
We then use the grammar to generate over one million complex question-SQL pairs given about 400k tables in different domains.

To pre-train \ty{should use pre-train instead of fine-tune, otherwise would be confusing. many msr people told me that} BERT for semantic parsing tasks, instead of using the masked language model (MLM) objective for free text, we propose a simple but efficient column objective.\vic{Not quite an ``instead of'' relationship here. The ``column objective'' is more like an intermediate task on top of MLM.}
Specifically, we reformulate the text-to-SQL problem as a classification task for each column, where the task is to predict the logical operator associated with that column in the question text.
As shown in Figure \ref{fig:tablebert}, above the BERT output for each column, we predict if any operations, such as aggregation, {\small \tt SELECT}, or {\small \tt WHERE}, were associated with it.
We show in Table \ref{tab:res_wikisql} that using the objective to pre-train BERT on the training data to be used later, before any training of the baseline model is done, can significantly improve the model performance.
In this manner, model training is effectively supervised by the exact same data twice in different ways. \ty{need to shorten and modify, skip some parts}

We then pre-train BERT on the augmented question-SQL pairs using the column objective to get \tablebert{}.
To test its effectiveness, we compare \tablebert{} and BERT \footnote{\roberta{} is finally used in all experiments shown in the paper. More details can be found in Section \ref{robert_ft}} about the four semantic parsing tasks under different settings.
\tablebert{} consistently outperforms BERT on all four tasks.
Our analysis shows that \tablebert{} can capture more examples that require harder schema mapping and contain more implicit and complex logic.
Our contributions can be summarized as follows:

\vspace{-2mm}
\begin{itemize}
    \item We propose a BERT fine-tuning objective that can significantly improve the performance of existing strong base models. 
    Also, we show that the simple training objective can efficiently capture schema linking and logic encoding.
    \item We develop an SCFG and automatically generate one million complex question-SQL pairs that covers complex and diverse SQL logic given about 400k tables. We then fine tune BERT on the augmented question-SQL examples to produce \tablebert{}. We show that \tablebert{} can significantly outperform BERT on four table-related semantic parsing tasks.
\end{itemize}

\ty{
introduction rewriting: 1) Like other table pre-training papers, first briefly rephrase and summarize the current introduction as our general motivation of this work. 2) then introduce recent prior pre-training work (tabert and tapas) for table-parsing, what methods they use and how effective they are. 3) move to our method by comparing with them, first briefly talk about our method, what are the differences, why prior work misses these differences, why they are important. mention specific motivations of our proposed pre-training method. 4) talk about the challenges of this ideas and prior attempts. e.g. overfit to canonical questions and hurt bert's original encoding ability. then introduce mlm loss and short training time to balance this. 5) experimented with multiple tasks with different settings to demonstrate its effectiveness and our hypothesises. 6) summarize our contributions
}
}

\section{Methodology}
\label{sec:methods}


\subsection{Motivation}
\label{ssec:mot}

Semantic parsing data is compositional because utterances are usually related to some formal representations such as logic forms and SQL queries.
Numerous prior works \citep{Berant14,wang2015parser,jia2016,iyer17,andreas-2020-good} have demonstrated the benefits of augmenting data using context-free grammar.
The augmented examples can be used to teach the model to generalize beyond the given training examples.

However, data augmentation becomes more complex and less beneficial if we want to apply it to generate data for a random domain.
More and more work~\citep{zhang19,herzig2020tapas,Campagna2020,Zhong2020GroundedAF} shows utilizing augmented data doesn’t always result in a significant performance gain in cross-domain semantic parsing end tasks.
The most likely reason for this is that models tend to overfit to the canonical input distribution especially the generated utterances are very different compared with the original ones.

Moreover, instead of directly training semantic parsers on the augmented data, our paper is the first to use the synthetic examples in pre-training in order to inject a compositional inductive bias to LMs and show it actually works if the overfitting problem is carefully addressed. 
To address the overfitting problem, in Section \ref{ssec:tabq}, we also include a small set of table related utterances in our pre-training data.
We add an MLM loss on them as a regularization factor, which requires the model to balance between real and synthetic examples during the pre-training.
We note that this consistently improves the performance on all downstream semantic parsing tasks (see Section \ref{sec:result}).
Finally, our pre-training method is much more data-efficient and save much more computational power than other prior work (Section \ref{pre-train-time}).

\subsection{Data Synthesis with Synchronous Context-Free Grammar}
\label{ssec:scfg}

We follow \citet{jia2016} to design our SCFG and apply it on a large amount of tables to populate new examples.
For example, as shown in Figure \ref{fig:intro_example}, by replacing substitutable column mentions (``locations''), table mentions (``performance''), values (``two''), and SQL logic phrases (``at least'') with the other possible candidates in the same group, our grammar generates new synthetic text-to-SQL examples with the same underlying SQL logic template.
We then pre-train BERT on the augmented examples to force it to discover substitutable fragments and learn the underlying logic template so that it is able to generalize to other similar questions. 
Meanwhile, BERT also benefits from pre-training on a large number of different columns, table names, and values in the generated data, which could potentially improve schema linking in semantic parsing tasks.

\begin{table*}[ht!]
\centering
\scalebox{0.73}{
\begin{tabular}{p{80mm}p{100mm}}
    \Xhline{2\arrayrulewidth}
    Non-terminals & Production rules  \\ \hline
    \textsc{Table} $\rightarrow t_i$
    \newline \textsc{Column} $\rightarrow c_i$ 
    \newline \textsc{value} $\rightarrow v_i$ 
    \newline \textsc{Agg} $\rightarrow$ $\langle$ \textsc{max}, \textsc{min}, \textsc{count}, \textsc{avg}, \textsc{sum}$\rangle$
    \newline \textsc{op} $\rightarrow$ $\langle$ $=$, $\leq$, $\neq$, ... , \textsc{like}, \textsc{between} $\rangle$ 
    \newline \textsc{sc} $\rightarrow$ $\langle$ \textsc{asc}, \textsc{desc} $\rangle$
    \newline \textsc{max} $\rightarrow$ $\langle$``maximum", ``the largest"...$\rangle$ 
    \newline \textsc{$\leq$} $\rightarrow$ $\langle$``no more than", ``no above"...$\rangle$ 
    \newline ... & 
    1. \textsc{root} $\rightarrow$ \big \langle ``For each \textsc{column0} , return how many times \textsc{table0} with \textsc{column1} \textsc{op0} \textsc{value0} ?", 
    \newline {\small \tt SELECT COLUMN0 , COUNT ( * ) WHERE COLUMN1 OP0 VALUE0 GROUP BY COLUMN0} \big \rangle
    \vspace{2mm}
    \newline 
    2. \textsc{root} $\rightarrow$ \big \langle ``What are the \textsc{column0} and \textsc{column1} of the \textsc{table0} whose \textsc{column2} is \textsc{op0} \textsc{agg0} \textsc{column2} ?",
    \newline {\small \tt SELECT COLUMN0 , COLUMN1 WHERE COLUMN2 {OP0} ( SELECT AGG0 ( COLUMN2 ) )} \big \rangle \\
    \Xhline{2\arrayrulewidth}
\end{tabular}}
\caption{Examples of non-terminals and production rules in our SCFG. Each production rule \textsc{root} $\rightarrow$ $\langle \alpha, \beta \rangle$ is built from some $(x, y) \in \mathcal{D}$ by replacing all terminal phrases with non-terminals. $t_i$, $c_i$, and $v_i$ stand for any table name, column name, entry value respectively. 
} 
\label{tb:scfg}
\vspace{-2mm}
\end{table*}

\paragraph{Grammar induction} 
To induce a cross-domain SCFG, we study examples in \spider{} since it is a publicly available dataset that includes the largest number of examples with complex compositionalities in different domains.
To further show the generality of our approach, we do not develop different SCFG for each downstream task.
Given a set of $(x, y)$ pairs in \spider{}, where $x$ and $y$ are the utterance and SQL query respectively.
We first define a set of non-terminal symbols for table names, column names, cell values, operations, etc.
For example, in Table~\ref{tb:scfg}, we group aggregation operations such as \textsc{max} as a non-terminal \textsc{agg}.
We can also replace the entities/phrases with their non-terminal types in SQL query to generate a SQL production rule $\beta$.
Then, we group $(x, y)$ pairs by similar SQL production rule $\beta$.
We automatically group and count Spider training examples by program templates, and select about 90 most frequent program templates $\beta$.
For each program template in the grammar, we randomly select roughly 4 corresponding natural language questions, manually replace entities/phrases with their corresponding non-terminal types to create natural language templates $\alpha$, and finally align them to generate each production rule \textsc{root} $\rightarrow$ $\langle \alpha, \beta \rangle$.
The manual alignment approximately takes a few hours.
About 500 \spider{} examples are studied to induce the SCFG.

\paragraph{Data augmentation} 
With $\langle \alpha, \beta \rangle$ pairs, we can simultaneously generate pseudo natural questions and corresponding SQL queries given a new table or database.
We first sample a production rule, and replace its non-terminals with one of corresponding terminals.
For example, we can map the non-terminal \textsc{agg} to \textsc{max} and ``maximum'' for the SQL query and the natural language sentence, respectively.
Also, table content is used in synthesizing our pre-training data. 
For example, if the sampled production rule contains a value (e.g., \textsc{value0}), we sample a value for the selected column from the table content and add it to the SQL and question templates. 
This way during pre-training, \tablebert{} can access the table content and learn the linking between values and columns.

We use \textsc{WikiTables}~\citep{Bhagavatula2015TabELEL}, which contains 1.6 million high-quality relational Wikipedia tables.
We remove tables with exactly the same column names and get about 340k tables and generate 413k question-SQL pairs given these tables.
Also, we generate another 62k question-SQL pairs using tables and databases in the training sets of \spider{} and \wikisql{}.
In total, our final pre-training dataset includes 475k question-SQL examples. 

We note that SCFG is usually crude~\citep{andreas-2020-good} especially when it is applied to augment data for different domains. 
In this work we don’t focus on how to develop a better SCFG that generates more natural utterances. We see this as a very interesting future work to explore.
Despite the fact that the SCFG is crude, our downstream task experiments show that it could be quite effective if some pre-training strategies are applied.


\subsection{Table Related Utterances}
\label{ssec:tabq}
As discussed in Section \ref{ssec:mot}, \tablebert{} is also pre-trained on human annotated questions over tables with a MLM objective.
We collected seven high quality datasets for textual-tabular data understanding (Table~\ref{tb:agg_data} in the Appendix), 
all of them contain Wikipedia tables or databases and the corresponding natural language utterances written by humans.
We only use tables and contexts as a pre-training resource and discard all the other human labels such as answers and SQL queries.

\subsection{Pre-Training \tablebert{}}
\label{robert_ft}

Unlike all the previous work where augmented data is used in the end task training, we apply the framework to language model pre-training.
Training semantic parsers is usually slow, and augmenting a large amount of syntactic pairs directly to the end task training data can be prohibitively slow or expensive.
In our work, we formulate text-to-SQL as a multi-class classification task for each column, which can be naturally combined with the MLM objective to pre-train BERT for semantic parsing.
Moreover, in this way, the learned knowledge can be easily and efficiently transferred to downstream semantic parsing tasks in the exact same way as BERT (shown in Section \ref{sec:result}).

\tablebert{} is initialized by RoBERTa$_{\textsc{large}}$~\citep{Liu2019RoBERTaAR} and further pre-trained on the synthetic data with SQL semantic loss and table-related data with MLM loss.
As shown in Figure \ref{fig:intro_example}, we follow \citet{Hwang2019ACE} to concatenate a user utterance and the column headers into a single flat sequence separated by the {\small \tt </s>}  token.
The user utterance can be either one of the original human utterances collected from the aggregated datasets or the canonical sentences sampled from the SCFG.
We add the table name at the beginning of each column if there are some complex schema inputs involving multiple tables.
We employ two objective functions for language model pre-training: 
1) masked-language modelling (MLM), and 
2) SQL semantic prediction (SSP).

\paragraph{MLM objective}
Intuitively, we would like to have a self-attention mechanism between natural language and table headers.
We conduct masking for both natural language sentence and table headers.
A small part of the input sequence is first replaced with the special token {\small \tt <mask>}.
The MLM loss is then computed by the cross-entropy function on predicting the masked tokens. 
We follow the default hyperparameters from \citet{Devlin2019BERTPO} with a 15\% masking probability. 

\paragraph{SSP objective}
With our synthetic natural language sentence and SQL query pairs, we can add an auxiliary task to train our column representations.
The proposed task is, given a natural language sentence and table headers, to predict whether a column appears in the SQL query and what operation is triggered.
We then convert all SQL sequence labels into operation classification labels for each column.
For example in the Figure~\ref{fig:intro_example}, the operation classification label of the column ``locations'' is \textsc{select and group by having}.
In total, there are 254 potential classes for operations in our experiments.

For a column or table indexed by $i$, we use the encoding of the special token {\small \tt </s>} right before it as its representation,
denoted as $\mathbf{x_i}$ to predict its corresponding operations.
On top of such representations, we apply a two-layer feed-forward network followed by a GELU activation layer~\citep{Hendrycks2016ABF} and a normalization layer~\citep{Ba2016LayerN} to the output representations. 
Formally, we compute the final vector representation of each column $\mathbf{y_i}$ by:
\begin{gather*}
    \mathbf{h} = \text{LayerNorm}(\text{GELU}(W_1\cdot\mathbf{x_i}))\\
    \mathbf{y_i} = \text{LayerNorm}(\text{GELU}(W_2 \cdot \mathbf{h}))
\end{gather*}

Finally, $\mathbf{y_i}$ is employed to compute the cross-entropy loss through a 
classification layer.
We sum losses from all columns in each training example for back-propagation.
For samples from the aggregated datasets, we only compute the MLM loss to update our model.
For samples from the synthetic data we generated, we compute only SSP loss to update our model.
More specifically, we mix 391k natural language utterances and 475k synthetic examples together as the final pre-training data. 
The examples in these two groups are randomly sampled during the pre-training, and MLM loss is computed if the selected example is a natural language question, otherwise SSP for a synthetic example.

\hide{
In this section, we first summarize the pre-existing datasets for textual-tabular data understanding used to pre-train our system. Then we introduce the proposed method to create natural language query and SQL labels. Finally, we present our pre-training strategy for table parsing.

\subsection{Data Aggregation}
\ty{for story telling, to move this part after the cfg part}
As shown in Table~\ref{tb:agg_data} in the Appendix, we collected seven high quality datasets for textual-tabular data understanding, including TabFact~\citep{chen2019tabfact}, LogicNLG~\citep{chen2020logical}, HybridQA~\citep{chen2020hybridqa}, WikiSQL~\citep{Zhong2017}, WikiTableQuestions~\citep{pasupat2015compositional}, ToTTo~\citep{parikh2020totto}, and Spider~\citep{Yu18emnlp}. 
All of them contain Wikipedia tables or databases and the corresponding natural language utterances written by humans.
Some utterances are questions over tables or databases, and some of them are descriptions of data content.
We only use tables and contexts as a pre-training resource and discard all the other human labels such as answers and SQL queries
\todo{Expect reviewer questions: why not use the provided SQL queries given we synthesized SQL ourselves}.

\subsection{Data Augmentation with Synchronous Context-Free Grammar}

Data augmentation for semantic parsing~\citep{Berant14,wang2015parser,jia2016,iyer17,Yu18syntax} makes an assumption, known as \hl{X context-free grammar (SCFG)}, 
\drr{this sentence is unclear.} that the meaning of semantic phrases is conditionally independent on the rest of the sentence. For example, column names and entry values such as ``European'' and ``countries'' are independent to the phrase ``at least 3 times''. Although this assumption is not always correct, especially in a cross-domain setting, it holds in most cases for generating canonical or natural language-like questions.


\begin{table*}[ht!]
\centering
\scalebox{0.73}{
\begin{tabular}{p{80mm}p{100mm}}
    \Xhline{2\arrayrulewidth}
    Non-terminals & Production rules  \\ \hline
    \textsc{Table} $\rightarrow t_i$
    \newline \textsc{Column} $\rightarrow c_i$ 
    \newline \textsc{value} $\rightarrow v_i$ 
    \newline \textsc{Agg} $\rightarrow$ $\langle$ \textsc{max}, \textsc{min}, \textsc{count}, \textsc{avg}, \textsc{sum}$\rangle$
    \newline \textsc{op} $\rightarrow$ $\langle$ $=$, $\leq$, $\neq$, ... , \textsc{like}, \textsc{between} $\rangle$ 
    \newline \textsc{sc} $\rightarrow$ $\langle$ \textsc{asc}, \textsc{desc} $\rangle$
    \newline \textsc{max} $\rightarrow$ $\langle$``maximum", ``the largest"...$\rangle$ 
    \newline \textsc{$\leq$} $\rightarrow$ $\langle$``no more than", ``no above"...$\rangle$ 
    \newline ... & 
    1. \textsc{root} $\rightarrow$ \big \langle ``For each \textsc{column0} , return how many times \textsc{table0} with \textsc{column1} \textsc{op0} \textsc{value0} ?", 
    \newline {\small \tt SELECT COLUMN0 , COUNT ( * ) WHERE COLUMN1 OP0 VALUE0 GROUP BY COLUMN0} \big \rangle
    \vspace{2mm}
    \newline 
    2. \textsc{root} $\rightarrow$ \big \langle ``What are the \textsc{column0} and \textsc{column1} of the \textsc{table0} whose \textsc{column2} is \textsc{op0} \textsc{agg0} \textsc{column2} ?",
    \newline {\small \tt SELECT COLUMN0 , COLUMN1 WHERE COLUMN2 {OP0} ( SELECT AGG0 ( COLUMN2 ) )} \big \rangle \\
    \Xhline{2\arrayrulewidth}
\end{tabular}}
\caption{Examples of non-terminals and production rules in our SCFG. Each production rule \textsc{root} $\rightarrow$ $\langle \alpha, \beta \rangle$ is built from some $(x, y) \in \mathcal{D}$ by replacing all terminal phrases with non-terminals. $t_i$, $c_i$, and $v_i$ stand for any table name, column name, entry value respectively. 
} 
\label{tb:scfg}
\vspace{-2mm}
\end{table*}


\paragraph{Grammar induction} Given a text-to-SQL dataset (e.g., \spider{}), we have a set of $(x, y)$ pairs, where $x$ and $y$ are the utterance and SQL query respectively.
We first define a set of non-terminal symbols for table names, column names, cell values, operations, etc.
As shown in Table~\ref{tb:scfg}, we group operations such as {\textsc{max}, \textsc{min}, \textsc{count}, \textsc{avg}, \textsc{sum}} as a non-terminals called \textsc{agg}.
We also replace the entities/phrases with their non-terminal types in SQL query to generate a SQL production rule $\beta$.
Then, we group $(x, y)$ pairs by similar SQL production rule $\beta$. 
We select the 90 most frequent $\beta$ and randomly select roughly 4 samples each, and manually align entities/phrases with their corresponding non-terminal types to collect natural language templates, $\alpha$. 


With $\langle \alpha, \beta \rangle$ pairs, we can simultaneously generate pseudo natural questions and corresponding SQL queries given a new table or database.
We first sample a production rule, and replace its non-terminals with one of corresponding terminals.
For example, we can map the non-terminal \textsc{agg} to \textsc{max} and ``maximum'' for the SQL query and the natural language sentence, respectively.




We use \textsc{WikiTables} \citep{Bhagavatula2015TabELEL}, which contains 1.6 million high-quality relational Wikipedia tables.
We remove tables with exactly the same column names and get about 340k tables and generate 413k question-SQL pairs given these tables.
Also, we generate another 62k question-SQL pairs using tables and databases in the training sets of \spider{} and \wikisql{}.
In total, our final pre-training dataset includes 475k question-SQL examples. 

\begin{figure*}[t!]
    \centering
    \includegraphics[width=0.98 \textwidth]{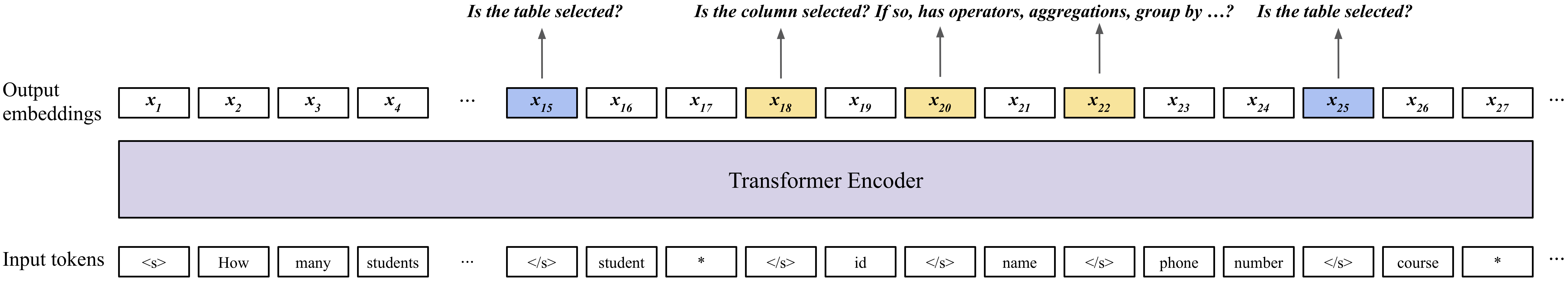}
    \caption{An illustration of the \tablebert column fine-tuning objective. 
    }
\label{fig:tablebert}
\end{figure*}

\ty{training data rewriting: 1) maybe first mention synthetic text-sql pair generation via CFG grammar and then mention, in order to leverage the augmented data in pretraining, it is very important to prevent bert from overfitting to unnatural questions. because of this, we add table-related data and mlm loss. 2) add a figure about the algorithm used to generate the text-sql pairs. list all details of the SCFG in table 3. 3) mention why use SQL as the semantic representation in pretraining (e.g. it is widely used to query structure data/dbs) and why use Spider to collect patterns in the grammar (e.g. spider contains examples with more diverse and complex semantics) 4) mention that the main goal of this paper is to demonstrate the augmented data could be used in bert pretraining and improving downstream tasks for table parsing. we don't focus on how to make the generated canonical questions look more natural, which this could a very interesting future work to explore to see if better generated questions can further improve the performance. 
}

\subsection{Pre-Training \tablebert{}}
\label{robert_ft}

\tablebert{} is initialized by RoBERTa$_{\textsc{large}}$~\citep{Liu2019RoBERTaAR} and further pre-train using table-related data.
We follow \citet{Hwang2019ACE} to concatenate a user utterance and the column headers into a single sequence separated by the {\small \tt </s>}  token.
The user utterance can be either one of the original human utterances $x$ collected from the aggregated datasets or the natural language sentences $\alpha$ we sampled from the production rules.
We add the table name at the beginning of each column if there are some complex schema inputs involving multiple tables.

We employ two objective functions for language model pre-training: 
1) masked-language modelling (MLM), and 
2) SQL-based operation prediction (SSP).

\paragraph{MLM objective}
Intuitively, we would like to have a self-attention mechanism between natural language and table headers.
We conduct masking for both natural language sentence and table headers. 
A small part of the input sequence is first replaced with the special token {\small \tt <mask>}. 
The MLM loss is then computed by the cross-entropy function on predicting the masked tokens. 
We follow the default hyperparameters from \citep{Devlin2019BERTPO} with a 15\% masking probability. 

\paragraph{SSP objective}
With our synthetic natural language sentence and SQL query pairs, we can add an auxiliary task to train our column representations.
The proposed task is, given a natural language sentence and table headers, to predict whether a column appears in the SQL query and what operation is triggered.
We then convert all SQL sequence labels into operation classification labels for each column.
For example in the Figure~\ref{fig:intro_example}, the operation classification label of the column ``Tourney'' is \textsc{where}.
For columns that appear in a nested query, we append the corresponding nested keywords before each operation.
For instance, since the column ``nation'' is in the \textsc{intersect} nested sub-query, its labels are \textsc{intersect select} and \textsc{intersect group by}.
For each additional table name, we only predict if it is selected for constructing possible \textsc{join}s in the \textsc{from} clause.
\hl{In total, there are 31??? potential classes for operations in our experiments.}

We use the encoding of the special token {\small \tt </s>} right before each column or table name to its corresponding predict operations. 
We also apply a 2-layer feed-forward network followed by a GELU activation layer~\citep{Hendrycks2016ABF} and a normalization layer~\citep{Ba2016LayerN} to the output representations.
Formally, we compute the final vector representation of each column $\mathbf{y_i}$ by:
\begin{gather*}
    \mathbf{h} = \text{LayerNorm}(\text{GELU}(\newvec{W_1}\cdot\mathbf{x_i}))\\
    \mathbf{y_i} = \text{LayerNorm}(\text{GELU}(\newvec{W_2} \cdot \mathbf{h}))
\end{gather*}
The representation of the special token {\small \tt </s>} $\mathbf{y_i}$ for each column is then use to compute the cross-entropy loss.
We sum losses from all columns in each training example for back-propagation.

\paragraph{Overall Training}
For samples from the aggregated datasets, we only compute the MLM loss to update our model.
For samples from the synthetic data we generated, we compute both MLM and SSP losses to update our model.
\hl{The weight the labels ... (Tao please add details here)}
}

\section{Experiments}
\label{exp:experiment}
We conduct experiments on four \textit{cross-domain} table semantic parsing tasks, where generalizing to unseen tables/databases at test time is required. 
We experiment with two different settings of table semantic parsing, fully supervised and weakly supervised setting.
The data statistics and examples on each task are shown in
Table~\ref{tb:data_stats} and Table~\ref{tb:examples} in the Appendix respectively.

\begin{table*}[ht!]
\centering
\scalebox{0.80}{
\begin{tabular}{ccccc}
\Xhline{2\arrayrulewidth}
Task \& Dataset & \# Examples & Resource & Annotation & Cross-domain\\\hline
\textsc{Spider} \cite{Yu18emnlp} & 10,181  & database & SQL & \checkmark\\
Fully-sup. \textsc{WikiSQL} \cite{Zhong2017} & 80,654  & single table & SQL & \checkmark\\\hline
\textsc{WikiTableQuestions} \cite{pasupat2015compositional} & 2,2033 & single table & answer & \checkmark \\
Weakly-sup. \textsc{WikiSQL} \cite{Zhong2017} & 80,654  & single table & answer & \checkmark\\\hline
\Xhline{2\arrayrulewidth}
\end{tabular}}
\caption{Overview of four table-based semantic parsing and question answering datasets in fully-supervised (top) and weakly-supervised (bottom) setting used in this paper. More details in Section \ref{exp:experiment}}
\label{tb:data_stats}
\vspace{-3mm}
\end{table*}

\subsection{Supervised Semantic Parsing}
We first evaluate \tablebert{} on two supervised semantic parsing tasks.
In a supervised semantic parsing scenario, given a question and a table or database schema,
a model is expected to generate the corresponding program.

\paragraph{\spider{}}
\spider{} \cite{Yu18emnlp} is a large text-to-SQL dataset.
It consists of 10k complex question-query pairs where many of the SQL queries contain multiple SQL keywords.
It also includes 200 databases where multiple tables are joined via foreign keys.
For the baseline model, we use RAT-SQL + \bert{}~\cite{wang-etal-2020-rat} which is the state-of-the-art model according to the official leaderboard. 
We followed the official Spider evaluation to report set match accuracy.

\paragraph{Fully-sup. \textsc{WikiSQL}} 
\wikisql{} \cite{Zhong2017} is a collection of over 80k questions and SQL query pairs over 30k Wikipedia tables.
We use \cite{Guo2019}, a competitive model on \wikisql{} built on SQLova \cite{Hwang2019ACE}, as our base model.
We adapt the same set of hyperparameters including batch size and maximum input length as in \citet{Guo2019}.
For a fair comparison, we only consider single models without execution-guided decoding and report execution accuracy.


\subsection{Weakly-supervised Semantic Parsing}
We also consider weakly-supervised semantic parsing tasks, which are very different from SQL-guided learning in pre-training.
In this setting, a question and its corresponding answer are given, but the underlying meaning representation (e.g., SQL queries) are unknown.

\paragraph{\wikitable{}}
This dataset contains question-denotation pairs over single Wikipedia tables \cite{pasupat2015compositional}.
The questions involve a variety of operations such as comparisons, superlatives, and aggregations, where some of them are hard to answered by SQL queries.

We used the model proposed by \citet{wang2019} which is the state-of-the-art parser on this task.
This model is a two-stage approach that first predicts a partial ``abstract program" and then refines that program while modeling structured alignments with differential dynamic programming.
The original model uses GloVe~\cite{pennington14} as word embeddings. We modified their implementation to encode question and column names in the same way as we do in our fine-tuning method that uses \roberta{} and \tablebert{}.

\paragraph{Weakly-sup. \wikisql{}}
In the weakly-supervised setting of \wikisql{}, only the answers (i.e., execution results of SQL queries) are available.
We also employed the model proposed by \citet{wang2019} as our baseline for this task.
We made the same changes and use the same experiment settings as described in the previous section for \wikitable{}.

\subsection{Implementation of \tablebert{}}
For fine-tuning \roberta{}, we modify the code of \roberta{} implemented by \citet{Wolf2019HuggingFacesTS} and follow the hyperparameters for fine-tuning RoBERTa on \textsc{race} tasks and use batch size 24, learning rate $1e$-5, and the Adam optimizer \cite{Kingma2014AdamAM}. 
We fine-tune \tablebert{} for 300k steps on eight 16GB Nvidia V100 GPUs.
The pre-training procedure can be done in less than 10 hours.
For all downstream experiments using \tablebert{} or \roberta{}, we always use a \bert{} specific optimizer to fine-tune them with a learning rate of $1e$-5, while using a model-specific optimizer with the respective learning rate for the rest of the base models.


\section{Experimental Results}
\label{sec:result}

We conducted experiments to answer the following two questions: 
1) Can \tablebert{} provide better representations for table semantic parsing tasks? 
2) What is the benefit of two pre-training objectives, namely MLM and SSP?
Since \tablebert{} is initialized by \roberta{}, we answer the first question by directly comparing the performance of base parser augmented with \tablebert{} and \roberta{} on table semantic parsing tasks. 
For the second question, we report the performance of \tablebert{} trained with MLM, SSP and also a variant with both of them (MLM+SSP).

\paragraph{Overall results} We report results on the four aforementioned tasks in Tables~ \ref{tab:res_spider}, \ref{tab:res_wikisql}, \ref{tab:res_wikitable}, and \ref{tab:res_wsp_wikisql} respectively.
Overall, base models augmented with \tablebert{} significantly outperforms the ones with \roberta{} by 3.7\% on \spider{}, 1.8\% on \wikitable{}, and 2.4\% on weakly-sup. \wikisql{}, and achieve new state-of-the-art results across all four tasks.
In most cases, the combined objective of MLM+SSP helps \tablebert{} achieve better performance when compared with independently using MLM and SSP.
Moreover, on the low-resource setting, \tablebert{} outperforms \roberta{} by 3.0\% in fully-sup. \wikisql{} and 3.9\% in \wikitable{}. 
Detailed results for each task are discussed as follows.

\begin{table*}[t!]
\centering
\adjustbox{valign=t}{\begin{minipage}[t]{0.52\textwidth}
    \centering
    \resizebox{\linewidth}{!}{
    \begin{tabular}{lcc}
    \Xhline{2\arrayrulewidth}
    Models & Dev. & Test \\ \hline
    Global-GNN \citep{Bogin2019GlobalRO} & 52.7 & 47.4 \\
    EditSQL \citep{zhang19} & 57.6 & 53.4\\
    IRNet \citep{Guo2019TowardsCT} & 61.9 & 54.7\\
    RYANSQL \citep{Choi2020RYANSQLRA} & 70.6 & 60.6\\
    TranX \citep{yin-etal-2020-tabert} & 64.5 & - \\
    \Xhline{2\arrayrulewidth}
    RAT-SQL  \citep{wang2019} & 62.7 & 57.2 \\
    \hspace{0.1cm} $w.$ BERT-large & 69.7 & 65.6 \\
    \hspace{0.1cm} $w.$ RoBERTa-large & 69.6 
    & - \\
    \hspace{0.1cm} $w.$ \tablebert{} (MLM) & 71.1(+1.4) 
    & - \\
    \hspace{0.1cm} $w.$ \tablebert{} (SSP) & \bf{73.6(+3.9)} 
    & 67.7(+2.1) \\
    \hspace{0.1cm} $w.$ \tablebert{} (MLM+SSP) & \bf{73.4(+3.7)} 
    & \bf{69.6(+4.0)} \\
    \Xhline{2\arrayrulewidth}
    \end{tabular}
    }
    \caption{Performance on \textsc{Spider}. We run each model three times by varying random seeds, and the average scores are shown.}
    \label{tab:res_spider}
\end{minipage}}%
\hfill
\adjustbox{valign=t}{\begin{minipage}[t]{0.46\textwidth}
    \centering
    \resizebox{\linewidth}{!}{
    \begin{tabular}{lcc}
    \Xhline{2\arrayrulewidth}
    Models & Dev. & Test \\\hline
    \citep{dong18} & 79.0 & 78.5 \\ 
    \citep{shi2018incsql} & 84.0 & 83.7 \\
    \citep{Hwang2019ACE} & 87.2 & 86.2 \\
    \citep{He2019XSQLRS} & 89.5 & 88.7 \\
    \citep{lyu2020hybrid} & 89.1 & 89.2 \\
    \Xhline{2\arrayrulewidth}
    \citep{Guo2019} & 90.3 & 89.2 \\
    \hspace{0.1cm} $w.$ RoBERTa-large & 91.2 & 90.6 \\
    \hspace{0.1cm} $w.$ \tablebert{} (MLM) & 91.4 & 90.7 \\
    \hspace{0.1cm} $w.$ \tablebert{} (SSP) & 91.2 & 90.7 \\
    \hspace{0.1cm} $w.$ \tablebert{} (MLM+SSP) & 91.2 & \bf{90.8} \\
    \hline
    \hspace{0.1cm} $w.$ RoBERTa-large (10k) & 79.6 & 79.2 \\
    \hspace{0.1cm} $w.$ \tablebert{} (MLM+SSP) (10k) & \bf{82.3(+2.7)} & \bf{82.2(+3.0)} \\
    \Xhline{2\arrayrulewidth}
    \end{tabular}
    }
    \caption{Performance on fully-sup. \textsc{WikiSQL}. All results are on execution accuracy without execution-guided decoding. }
    \label{tab:res_wikisql}
\end{minipage}}%
\centering
\label{fig:compare_fig}
\end{table*}


\paragraph{\spider{}} Results on \spider{} are shown in Table~\ref{tab:res_spider}.
 When augmented with \tablebert{}, the model achieves significantly better performance compared with the baselines using BERT and \roberta{}.
Our best model, \tablebert{} with MLM+SSP achieves the new state-of-the-art performance, surpassing previous one (RAT-SQL+BERT-large) by a margin of 4\%. 
Notably, most previous top systems use pre-trained contextual representations (e.g., BERT, TaBERT), indicating the importance of 
such representations for the cross-domain parsing task.  


\paragraph{Fully sup. \wikisql{}} 
Results on \wikisql{} are shown in Table~\ref{tab:res_wikisql}.
All \tablebert{} models achieve nearly the same performance as \roberta{}. We suspect it is the relatively large training size and easy SQL pattern of \wikisql{} make the improvement hard, comparing to \spider{}. Hence, we set up a low-resource setting where we only use 10k examples from the training data. As shown in the bottom two lines of Table~\ref{tab:res_wikisql},
\tablebert{} improves the performance of the SQLova model by 3.0\% compared to \roberta{}, indicating that \tablebert{} can make the base parser more sample-efficient. 


\paragraph{\wikitable{}}
Results on \wikitable{} are shown in Table~\ref{tab:res_wikitable}.
By using \roberta{} and \tablebert{} to encode question and column inputs, the performance of \citet{wang2019} can be boosted significantly (	$>$6\%). 
Compared with \roberta{}, our best model with \tablebert{} (MLM+SSP) can further improve the performance by 1.8\%, leading to a new state-of-the-art performance on this task.
Similar to the low-resource experiments for \wikisql{}, we also show the performance of the model when trained with only 10\% of the training data. As shown at the bottom two lines Table~\ref{tab:res_wikitable}, \tablebert{} (MLM + SSP) obtains much better performance than \roberta{}, again showing its superiority of providing better representations. 

\paragraph{Weakly sup. \wikisql{}}
Results on weakly supervised \wikisql{} are shown in Table~\ref{tab:res_wsp_wikisql}.
\tablebert{} with MLM+SSP again achieves the best performance when compared with other baselines, obtain the new state-of-the-art results of 84.7\% on this task. 
It is worth noting that our best model here  
is also better than many models trained in the fully-supervised setting in Table \ref{tab:res_wikisql}. This suggests that inductive biases injected in pre-trained representation of \tablebert{} can significantly help combat the issue of spurious programs introduced by learning from denotations~\cite{pasupat2015compositional,wang2019} when gold programs are not available.

\begin{table*}[t!]
\centering
\begin{minipage}[t]{0.52\textwidth}
    \centering
    \resizebox{\linewidth}{!}{
    \begin{tabular}{lcc}
    \Xhline{2\arrayrulewidth}
    Models & Dev. & Test \\\hline
    \citep{Liang2018MemoryAP} & 42.3 & 43.1 \\ 
    \citep{Dasigi2019IterativeSF} & 42.1 & 43.9 \\ 
    \citep{Agarwal2019LearningTG} & 43.2 & 44.1 \\ 
    \citep{herzig2020tapas} & - &48.8 \\
    \citep{yin2020tabert} & 52.2 & 51.8 \\
    \Xhline{2\arrayrulewidth}
    \citep{wang2019} & 43.7 & 44.5 \\
    \hspace{0.1cm} $w.$ RoBERTa-large & 50.7(+7.0) & 50.9(+6.4) \\
    \hspace{0.1cm} $w.$ \tablebert{} (MLM) & 51.5(+7.8) & 51.7(+7.2) \\
    \hspace{0.1cm} $w.$ \tablebert{} (SSP) & 51.2(+7.5) & 51.1(+6.6) \\
    \hspace{0.1cm} $w.$ \tablebert{} (MLM+SSP) & \textbf{51.9(+8.2)} & \textbf{52.7(+8.2)} \\
    \hline
    \hspace{0.1cm} $w.$ RoBERTa-large $\times$10\% & 37.3 & 38.1 \\
    \hspace{0.1cm} $w.$ \tablebert{} (MLM+SSP) $\times$10\% & \textbf{40.4(+3.1)} & \textbf{42.0(+3.9)}\\
    \Xhline{2\arrayrulewidth}
    \end{tabular}
    }
    \caption{Performance on \textsc{WikiTableQuestions}. Results trained on 10\% of the data are shown at the bottom.}
    \label{tab:res_wikitable}
\end{minipage}%
\hfill
\begin{minipage}[t]{0.465\textwidth}
    \centering
    \resizebox{\linewidth}{!}{
    \begin{tabular}{lcc}
    \Xhline{2\arrayrulewidth}
    Models & Dev. & Test \\\hline
    \citep{Liang2018MemoryAP} & 72.2 & 72.1 \\ 
    \citep{Agarwal2019LearningTG} & 74.9 & 74.8 \\
    \citep{min2019discrete} & 84.4 & 83.9 \\
    \citep{herzig2020tapas} & 85.1 & 83.6 \\
    \Xhline{1\arrayrulewidth}
    \citep{wang2019} & 79.4 & 79.3 \\
    \hspace{0.1cm} $w.$ RoBERTa-large & 82.3 (+2.9) & 82.3 (+3.0) \\
    \hspace{0.1cm} $w.$ \tablebert{} (MLM) & 83.3 (+3.9) & 83.5 (+4.2) \\
    \hspace{0.1cm} $w.$ \tablebert{} (SSP) & 83.5(+4.1) & 83.7 (+4.4) \\
    \hspace{0.1cm} $w.$ \tablebert{} (MLM+SSP) & \bf{85.9} (+6.5) & \bf{84.7} (+5.4) \\
    \Xhline{2\arrayrulewidth}
    \end{tabular}}
    \caption{Performance on weakly-sup. \textsc{WikiSQL}. We use \citep{wang2019} as our base model. 
    }
    \label{tab:res_wsp_wikisql}
    \end{minipage}%
\label{fig:compare_fig}
\end{table*}



\section{Analysis}

\paragraph{Pre-training objectives} 
\tablebert{} trained with both MLM and SSP loss consistently outperforms the one trained with one of them (MLM+SSP vs. MLM only or SSP only).
\tablebert{} (MLM) usually improves the performance by around 1\% such as 1.4\% gain on \spider{} (dev), 0.8\% on \wikitable{}, and 1.2\% on weakly-sup. \wikisql{}.
By pre-training on the synthetic text-to-SQL examples, \tablebert{} (SSP), we can see a similar performance gain on these tasks too except 3.9\% improvement on \spider{} dev, which is what we expected (grammar is overfitted to \spider{}). 
By pre-training with both MLM and SSP on the combined data, \tablebert{} (MLM+SSP) consistently and significantly outperforms the one pre-trained with MLM or SSP separately (e.g., about +2\% on Spider, +1.5\% on WikiTableQuestions, and +1.2\% on weakly-sup WikiSQL.). 
This contributes to our key argument in the paper: in order to effectively inject compositional inductive bias to LM, pre-training on synthetic data should be regularized properly (using SSP+MLM together instead of SSP or MLM only) in order to balance between preserving the original BERT encoding ability and injecting compositional inductive bias, otherwise, the improvements are not robust and limited (using SSP or MLM only).

\paragraph{Generalization}
As mentioned in Section \ref{ssec:scfg}, we design our SCFG solely based on \spider{}, and then sample from it to generate synthetic examples.
Despite the fact that \tablebert{} pre-trained on such corpus is optimized to the \spider{} data distribution, which is very different from \wikisql{} and \wikitable{}, \tablebert{} is still able to improve performance on the two datasets.
In particular, for \wikitable{} where the underlying distribution of programs (not necessarily in the form of SQL) are latent, \tablebert{} can still help a parser generalize better, indicating \tablebert{} can be beneficial for general table understanding even though it is pre-trained on SQL specific semantics. 
We believe that incorporating rules from a broader range of datasets (e.g. \wikitable{}) would further improve the performance.
However, in this paper, we study rules from only the \spider{} dataset and test the effectiveness on other unseen datasets with different different underlying rules on purpose in order to show the generality of our method.

Even though \tablebert{} is pre-trained on synthetic text-to-SQL data, the proposed pre-training method can also be applied to many other semantic parsing tasks with different formal programs (e.g., logic forms); and we also demonstrated the effectness of \tablebert{} on non text-to-SQL tasks (weakly-supervised \wikisql{} and \wikitable{} where no programs are used, training is supervised by only answers/cell values) the underlying distribution of programs (not necessarily in the form of SQL) are latent. 
Furthermore, to design the SCFG and synthesize data with the corresponding programs labeled, we can use any formal programs such as the logic form or SParQL, and then employ the data to pre-train GraPPa. 
In this paper we choose SQL as the formal program to represent the formal representation of the questions simply because more semantic parsing datasets are labeled in SQL.



\paragraph{Pre-training time and data}
\label{pre-train-time}
Our experiments on the \spider{} and \wikitable{} tasks show that longer pre-training doesn’t improve and can even hurt the performance of the pre-trained model. 
This also indicates that synthetic data should be carefully used in order to balance between preserving the original BERT encoding ability and injecting compositional inductive bias.
The best result on \spider{} is achieved by using \tablebert{} pre-trained for only 5 epochs on our relatively small pre-training dataset.
Compared to other recent pre-training methods for semantic parsing such as TaBERT \citep{yin-etal-2020-tabert} and TAPAS \citep{Herzig2020TAPASWS}, \tablebert{} achieves the state-of-the-art performance (incorporated with strong base systems) on the four representative table semantic parsing tasks \emph{in less 10 hours on only 8 16GB Nvidia V100 GPUs} (6 days on more than 100 V100 GPUs and 3 days on 32 TPUs for TaBERT and TAPAS respectively)
Moreover, we encourage future work on studying how the size and quality of synthetic data would affect the end task performance.
Also, \tablebert{} (MLM+SSP) consistently outperforms other settings, which indicates that using MLM on the human annotated data is important.

\paragraph{Pre-training vs. training data augmentation}
Many recent work \citep{zhang19,herzig2020tapas,Campagna2020,Zhong2020GroundedAF} in semantic parsing and dialog state tracking show that training models on a combination of the extra synthetic data and original training data does not improve or even hurt the performance.
For example, \citep{Zhong2020GroundedAF} synthesize data on training databases in several semantic parsing tasks including \spider{}, and find that training with this data augmentation leads to overfitting on the synthetic data and decreases the performance.
In contrast, our pre-training approach could effectively utilize a large amount of synthesized data and improve downstream task performance.
Also, the base parser with a \tablebert{} encoder could usually converge to a higher performance in shorter time (see Section \ref{sec:add_analysis}).

\section{Related Work}
\label{related_work}

\paragraph{Textual-tabular data understanding} Real-world data exist in both structured and unstructured forms. Recently the field has witnessed a surge of interest in joint textual-tabular data understanding problems, such as table semantic parsing~\citep{Zhong2017,Yu18emnlp}, question answering~\citep{pasupat2015compositional,chen2020hybridqa}, retrieval~\citep{Zhang2019tab}, fact-checking~\citep{chen2019tabfact} and summarization~\citep{parikh2020totto,radev2020dart}. 
While most work focus on single tables, often obtained from the Web, some have extended modeling to more complex structures such as relational databases~\citep{cathy18,Yu18emnlp,wang-etal-2020-rat}. 
All of these tasks can benefit from better representation of the input text and different components of the table, and most importantly, an effective contextualization across the two modalities. Our work aims at obtaining high-quality cross-modal representation via pre-training to potentially benefit all downstream tasks.

\paragraph{Pre-training for NLP tasks} \tablebert{} is inspired by recent advances in pre-training for text such as~\citep{Devlin2019BERTPO,Liu2019RoBERTaAR,lewis-etal-2020-bart,lewis2020pre,Guu2020REALMRL}. Seminal work in this area shows that textual representation trained using conditional language modeling objectives significantly improves performance on various downstream NLP tasks. This triggered an exciting line of research work under the themes of (1) cross-modal pre-training that involves text~\citep{DBLP:conf/nips/LuBPL19,peters-etal-2019-knowledge,yin-etal-2020-tabert,Herzig2020TAPASWS} and (2) pre-training architectures and objectives catering subsets of NLP tasks~\citep{lewis-etal-2020-bart,lewis2020pre,Guu2020REALMRL}. \tablebert{} extends these two directions further. The closest work to ours are TaBERT~\citep{yin-etal-2020-tabert} and \textsc{TaPas}~\citep{Herzig2020TAPASWS}. 
Both are trained over millions of web tables and relevant but noisy textual context.
In comparison,~\tablebert{} is pre-trained with a novel training objective, over synthetic data plus a much smaller but cleaner collection of text-table datasets. 

\paragraph{Data augmentation for semantic parsing} 
Our work was inspired by existing work on data augmentation for semantic parsing~\citep{Berant14,wang2015parser,jia2016,iyer17,Yu18syntax}.
\citet{Berant14} employed a rule-based approach to generate canonical natural language utterances given a logical form. 
A paraphrasing model was then used to choose the canonical utterance that best paraphrases the input and to output the corresponding logical form.
In contrast, \citet{jia2016} used prior knowledge in structural regularities to induce an SCFG and then directly use the grammar to generate more training data, which resulted in a significant improvement on the tasks.
Unlike these works which augment a relatively small number of data and use them directly in end task training, we synthesize a large number of texts with SQL logic grounding to each table cheaply 
and use them for pre-training.

\section{Conclusion and Future Work}
In this paper, we proposed a novel and effective pre-training approach for table semantic parsing.
We developed a context-free grammar to automatically generate a large amount of question-SQL pairs.
Then, we introduced \tablebert{}, which is an LM that is pre-trained on the synthetic examples with SQL semantic loss.
We discovered that, in order to better leverage augmented data, it is important to add MLM loss on a small amount of table related utterances.
Results on four semantic parsing tasks demonstrated that \tablebert{} significantly outperforms \roberta{}. 

While the pre-training method is surprisingly effective in its current form, we view these results primarily as an invitation for more future work in this direction.
For example, this work relies on a hand-crafted grammar which often generates unnatural questions; Further improvements are likely to be made by applying more sophisticated data augmentation techniques. 
Also, it would be interesting to study the relative impact of the two objectives (MLM and SSP) by varying the respective number of pre-training examples.
Furthermore, pre-training might benefit from synthesizing data from a more compositional grammar with a larger logical form coverage, and also from supervising by a more compositional semantic signals.

\bibliography{iclr2021_conference}

\begin{thebibliography}{56}
\providecommand{\natexlab}[1]{#1}
\providecommand{\url}[1]{\texttt{#1}}
\expandafter\ifx\csname urlstyle\endcsname\relax
  \providecommand{\doi}[1]{doi: #1}\else
  \providecommand{\doi}{doi: \begingroup \urlstyle{rm}\Url}\fi

\bibitem[Agarwal et~al.(2019)Agarwal, Liang, Schuurmans, and
  Norouzi]{Agarwal2019LearningTG}
Rishabh Agarwal, Chen Liang, Dale Schuurmans, and Mohammad Norouzi.
\newblock Learning to generalize from sparse and underspecified rewards.
\newblock In \emph{ICML}, 2019.

\bibitem[Andreas(2020)]{andreas-2020-good}
Jacob Andreas.
\newblock Good-enough compositional data augmentation.
\newblock In \emph{Proceedings of the 58th Annual Meeting of the Association
  for Computational Linguistics}, pp.\  7556--7566, Online, July 2020.
  Association for Computational Linguistics.
\newblock \doi{10.18653/v1/2020.acl-main.676}.

\bibitem[Artzi \& Zettlemoyer(2013)Artzi and Zettlemoyer]{artzi13}
Yoav Artzi and Luke Zettlemoyer.
\newblock Weakly supervised learning of semantic parsers for mapping
  instructions to actions.
\newblock \emph{Transactions of the Association forComputational Linguistics},
  2013.

\bibitem[Ba et~al.(2016)Ba, Kiros, and Hinton]{Ba2016LayerN}
Jimmy Ba, Jamie~Ryan Kiros, and Geoffrey~E. Hinton.
\newblock Layer normalization.
\newblock \emph{ArXiv}, abs/1607.06450, 2016.

\bibitem[Berant \& Liang(2014)Berant and Liang]{Berant14}
Jonathan Berant and Percy Liang.
\newblock Semantic parsing via paraphrasing.
\newblock In \emph{Proceedings of the 52nd Annual Meeting of the Association
  for Computational Linguistics (Volume 1: Long Papers)}, pp.\  1415--1425,
  Baltimore, Maryland, June 2014. Association for Computational Linguistics.

\bibitem[Bhagavatula et~al.(2015)Bhagavatula, Noraset, and
  Downey]{Bhagavatula2015TabELEL}
Chandra Bhagavatula, Thanapon Noraset, and Doug Downey.
\newblock Tabel: Entity linking in web tables.
\newblock In \emph{International Semantic Web Conference}, 2015.

\bibitem[Bogin et~al.(2019)Bogin, Gardner, and Berant]{Bogin2019GlobalRO}
Ben Bogin, Matt Gardner, and Jonathan Berant.
\newblock Global reasoning over database structures for text-to-sql parsing.
\newblock \emph{ArXiv}, abs/1908.11214, 2019.

\bibitem[Campagna et~al.(2020)Campagna, Foryciarz, Moradshahi, and
  Lam]{Campagna2020}
Giovanni Campagna, Agata Foryciarz, Mehrad Moradshahi, and Monica~S. Lam.
\newblock Zero-shot transfer learning with synthesized data for multi-domain
  dialogue state tracking.
\newblock In \emph{Proceedings of 58th Annual Meeting of the Association for
  Computational Linguistics (Volume 1: Long Papers)}, 2020.

\bibitem[Chen et~al.(2019)Chen, Wang, Chen, Zhang, Wang, Li, Zhou, and
  Wang]{chen2019tabfact}
Wenhu Chen, Hongmin Wang, Jianshu Chen, Yunkai Zhang, Hong Wang, Shiyang Li,
  Xiyou Zhou, and William~Yang Wang.
\newblock Tabfact: A large-scale dataset for table-based fact verification.
\newblock \emph{arXiv preprint arXiv:1909.02164}, 2019.

\bibitem[Chen et~al.(2020)Chen, Zha, Chen, Xiong, Wang, and
  Wang]{chen2020hybridqa}
Wenhu Chen, Hanwen Zha, Zhiyu Chen, Wenhan Xiong, Hong Wang, and William Wang.
\newblock Hybridqa: A dataset of multi-hop question answering over tabular and
  textual data.
\newblock \emph{arXiv preprint arXiv:2004.07347}, 2020.

\bibitem[Choi et~al.(2020)Choi, Shin, Kim, and Shin]{Choi2020RYANSQLRA}
Donghyun Choi, Myeong~Cheol Shin, Eunggyun Kim, and Dong~Ryeol Shin.
\newblock Ryansql: Recursively applying sketch-based slot fillings for complex
  text-to-sql in cross-domain databases.
\newblock \emph{ArXiv}, abs/2004.03125, 2020.

\bibitem[Dasigi et~al.(2019)Dasigi, Gardner, Murty, Zettlemoyer, and
  Hovy]{Dasigi2019IterativeSF}
Pradeep Dasigi, Matt Gardner, Shikhar Murty, Luke~S. Zettlemoyer, and Eduard~H.
  Hovy.
\newblock Iterative search for weakly supervised semantic parsing.
\newblock In \emph{NAACL-HLT}, 2019.

\bibitem[Devlin et~al.(2019)Devlin, Chang, Lee, and
  Toutanova]{Devlin2019BERTPO}
Jacob Devlin, Ming-Wei Chang, Kenton Lee, and Kristina Toutanova.
\newblock Bert: Pre-training of deep bidirectional transformers for language
  understanding.
\newblock In \emph{NAACL-HLT}, 2019.

\bibitem[Dong \& Lapata(2018)Dong and Lapata]{dong18}
Li~Dong and Mirella Lapata.
\newblock Coarse-to-fine decoding for neural semantic parsing.
\newblock In \emph{Proceedings of the 56th Annual Meeting of the Association
  for Computational Linguistics (Volume 1: Long Papers)}, pp.\  731--742.
  Association for Computational Linguistics, 2018.
\newblock URL \url{http://aclweb.org/anthology/P18-1068}.

\bibitem[Finegan-Dollak et~al.(2018)Finegan-Dollak, Kummerfeld, Zhang,
  Ramanathan, Sadasivam, Zhang, and Radev]{cathy18}
Catherine Finegan-Dollak, Jonathan~K. Kummerfeld, Li~Zhang, Karthik
  Ramanathan~Dhanalakshmi Ramanathan, Sesh Sadasivam, Rui Zhang, and Dragomir
  Radev.
\newblock Improving text-to-sql evaluation methodology.
\newblock In \emph{ACL 2018}. Association for Computational Linguistics, 2018.

\bibitem[Guo et~al.(2019)Guo, Zhan, Gao, Xiao, Lou, Liu, and
  Zhang]{Guo2019TowardsCT}
Jiaqi Guo, Zecheng Zhan, Yan Gao, Yan Xiao, Jian-Guang Lou, Ting Liu, and
  Dongmei Zhang.
\newblock Towards complex text-to-sql in cross-domain database with
  intermediate representation.
\newblock In \emph{ACL}, 2019.

\bibitem[Guo \& Gao(2019)Guo and Gao]{Guo2019}
Tong Guo and Huilin Gao.
\newblock Content enhanced bert-based text-to-sql generation.
\newblock Technical report, 2019.

\bibitem[Guu et~al.(2020)Guu, Lee, Tung, Pasupat, and Chang]{Guu2020REALMRL}
Kelvin Guu, Kenton Lee, Zora Tung, Panupong Pasupat, and Ming-Wei Chang.
\newblock Realm: Retrieval-augmented language model pre-training.
\newblock \emph{ArXiv}, abs/2002.08909, 2020.

\bibitem[He et~al.(2019)He, Mao, Chakrabarti, and Chen]{He2019XSQLRS}
Pengcheng He, Yi~Mao, Kaushik Chakrabarti, and Weizhu Chen.
\newblock X-sql: reinforce schema representation with context.
\newblock \emph{ArXiv}, abs/1908.08113, 2019.

\bibitem[Hendrycks \& Gimpel(2016)Hendrycks and Gimpel]{Hendrycks2016ABF}
Dan Hendrycks and Kevin Gimpel.
\newblock A baseline for detecting misclassified and out-of-distribution
  examples in neural networks.
\newblock \emph{ArXiv}, abs/1610.02136, 2016.

\bibitem[Herzig \& Berant(2018)Herzig and Berant]{Herzig2018}
Jonathan Herzig and Jonathan Berant.
\newblock Decoupling structure and lexicon for zero-shot semantic parsing.
\newblock In \emph{Proceedings of the 2018 Conference on Empirical Methods in
  Natural Language Processing}, pp.\  1619--1629. Association for Computational
  Linguistics, 2018.

\bibitem[Herzig et~al.(2020{\natexlab{a}})Herzig, Nowak, M{\"u}ller, Piccinno,
  and Eisenschlos]{Herzig2020TAPASWS}
Jonathan Herzig, P.~Nowak, Thomas M{\"u}ller, Francesco Piccinno, and
  Julian~Martin Eisenschlos.
\newblock Tapas: Weakly supervised table parsing via pre-training.
\newblock In \emph{ACL}, 2020{\natexlab{a}}.

\bibitem[Herzig et~al.(2020{\natexlab{b}})Herzig, Nowak, M{\"u}ller, Piccinno,
  and Eisenschlos]{herzig2020tapas}
Jonathan Herzig, Pawe{\l}~Krzysztof Nowak, Thomas M{\"u}ller, Francesco
  Piccinno, and Julian~Martin Eisenschlos.
\newblock Tapas: Weakly supervised table parsing via pre-training.
\newblock \emph{arXiv preprint arXiv:2004.02349}, 2020{\natexlab{b}}.

\bibitem[Hwang et~al.(2019)Hwang, Yim, Park, and Seo]{Hwang2019ACE}
Wonseok Hwang, Jinyeung Yim, Seunghyun Park, and Minjoon Seo.
\newblock A comprehensive exploration on wikisql with table-aware word
  contextualization.
\newblock \emph{ArXiv}, abs/1902.01069, 2019.

\bibitem[Iyer et~al.(2017)Iyer, Konstas, Cheung, Krishnamurthy, and
  Zettlemoyer]{iyer17}
Srinivasan Iyer, Ioannis Konstas, Alvin Cheung, Jayant Krishnamurthy, and Luke
  Zettlemoyer.
\newblock Learning a neural semantic parser from user feedback.
\newblock \emph{CoRR}, abs/1704.08760, 2017.

\bibitem[Jia \& Liang(2016)Jia and Liang]{jia2016}
Robin Jia and Percy Liang.
\newblock Data recombination for neural semantic parsing.
\newblock In \emph{Proceedings of the 54th Annual Meeting of the Association
  for Computational Linguistics (Volume 1: Long Papers)}, 2016.

\bibitem[Kingma \& Ba(2014)Kingma and Ba]{Kingma2014AdamAM}
Diederik~P. Kingma and Jimmy Ba.
\newblock Adam: A method for stochastic optimization.
\newblock \emph{CoRR}, abs/1412.6980, 2014.

\bibitem[Lewis et~al.(2020{\natexlab{a}})Lewis, Ghazvininejad, Ghosh,
  Aghajanyan, Wang, and Zettlemoyer]{lewis2020pre}
Mike Lewis, Marjan Ghazvininejad, Gargi Ghosh, Armen Aghajanyan, Sida Wang, and
  Luke Zettlemoyer.
\newblock Pre-training via paraphrasing.
\newblock \emph{arXiv preprint arXiv:2006.15020}, 2020{\natexlab{a}}.

\bibitem[Lewis et~al.(2020{\natexlab{b}})Lewis, Liu, Goyal, Ghazvininejad,
  Mohamed, Levy, Stoyanov, and Zettlemoyer]{lewis-etal-2020-bart}
Mike Lewis, Yinhan Liu, Naman Goyal, Marjan Ghazvininejad, Abdelrahman Mohamed,
  Omer Levy, Veselin Stoyanov, and Luke Zettlemoyer.
\newblock {BART}: Denoising sequence-to-sequence pre-training for natural
  language generation, translation, and comprehension.
\newblock In \emph{Proceedings of the 58th Annual Meeting of the Association
  for Computational Linguistics}, pp.\  7871--7880, Online, July
  2020{\natexlab{b}}. Association for Computational Linguistics.
\newblock \doi{10.18653/v1/2020.acl-main.703}.
\newblock URL \url{https://www.aclweb.org/anthology/2020.acl-main.703}.

\bibitem[Li \& Jagadish(2014)Li and Jagadish]{li2014constructing}
Fei Li and HV~Jagadish.
\newblock Constructing an interactive natural language interface for relational
  databases.
\newblock \emph{VLDB}, 2014.

\bibitem[Liang et~al.(2018)Liang, Norouzi, Berant, Le, and
  Lao]{Liang2018MemoryAP}
Chen Liang, Mohammad Norouzi, Jonathan Berant, Quoc~V. Le, and Ni~Lao.
\newblock Memory augmented policy optimization for program synthesis and
  semantic parsing.
\newblock In \emph{NeurIPS}, 2018.

\bibitem[Lin et~al.(2020)Lin, Socher, and Xiong]{LinRX2020:BRIDGE}
Xi~Victoria Lin, Richard Socher, and Caiming Xiong.
\newblock Bridging textual and tabular data for cross-domain text-to-sql
  semantic parsing.
\newblock In \emph{Findings of the 2020 Conference on Empirical Methods in
  Natural Language Processing, {EMNLP Findings} 2020, November 16th-20th,
  2020}, 2020.

\bibitem[Liu et~al.(2019)Liu, Ott, Goyal, Du, Joshi, Chen, Levy, Lewis,
  Zettlemoyer, and Stoyanov]{Liu2019RoBERTaAR}
Yinhan Liu, Myle Ott, Naman Goyal, Jingfei Du, Mandar Joshi, Danqi Chen, Omer
  Levy, Mike Lewis, Luke~S. Zettlemoyer, and Veselin Stoyanov.
\newblock Roberta: A robustly optimized bert pretraining approach.
\newblock \emph{ArXiv}, abs/1907.11692, 2019.

\bibitem[Lu et~al.(2019)Lu, Batra, Parikh, and Lee]{DBLP:conf/nips/LuBPL19}
Jiasen Lu, Dhruv Batra, Devi Parikh, and Stefan Lee.
\newblock Vilbert: Pretraining task-agnostic visiolinguistic representations
  for vision-and-language tasks.
\newblock In Hanna~M. Wallach, Hugo Larochelle, Alina Beygelzimer, Florence
  d'Alch{\'{e}}{-}Buc, Emily~B. Fox, and Roman Garnett (eds.), \emph{Advances
  in Neural Information Processing Systems 32: Annual Conference on Neural
  Information Processing Systems 2019, NeurIPS 2019, 8-14 December 2019,
  Vancouver, BC, Canada}, pp.\  13--23, 2019.

\bibitem[Lyu et~al.(2020)Lyu, Chakrabarti, Hathi, Kundu, Zhang, and
  Chen]{lyu2020hybrid}
Qin Lyu, Kaushik Chakrabarti, Shobhit Hathi, Souvik Kundu, Jianwen Zhang, and
  Zheng Chen.
\newblock Hybrid ranking network for text-to-sql.
\newblock Technical Report MSR-TR-2020-7, Microsoft Dynamics 365 AI, March
  2020.

\bibitem[Min et~al.(2019)Min, Chen, Hajishirzi, and
  Zettlemoyer]{min2019discrete}
Sewon Min, Danqi Chen, Hannaneh Hajishirzi, and Luke Zettlemoyer.
\newblock A discrete hard em approach for weakly supervised question answering.
\newblock In \emph{EMNLP}, 2019.

\bibitem[Parikh et~al.(2020)Parikh, Wang, Gehrmann, Faruqui, Dhingra, Yang, and
  Das]{parikh2020totto}
Ankur~P Parikh, Xuezhi Wang, Sebastian Gehrmann, Manaal Faruqui, Bhuwan
  Dhingra, Diyi Yang, and Dipanjan Das.
\newblock Totto: A controlled table-to-text generation dataset.
\newblock \emph{arXiv preprint arXiv:2004.14373}, 2020.

\bibitem[Pasupat \& Liang(2015)Pasupat and Liang]{pasupat2015compositional}
Panupong Pasupat and Percy Liang.
\newblock Compositional semantic parsing on semi-structured tables.
\newblock In \emph{Proceedings of the 53rd Annual Meeting of the Association
  for Computational Linguistics and the 7th International Joint Conference on
  Natural Language Processing of the Asian Federation of Natural Language
  Processing, {ACL} 2015, July 26-31, 2015, Beijing, China, Volume 1: Long
  Papers}, pp.\  1470--1480, 2015.

\bibitem[Pennington et~al.(2014)Pennington, Socher, and Manning]{pennington14}
Jeffrey Pennington, Richard Socher, and Christopher~D. Manning.
\newblock Glove: Global vectors for word representation.
\newblock In \emph{{EMNLP}}, pp.\  1532--1543. {ACL}, 2014.

\bibitem[Peters et~al.(2019)Peters, Neumann, Logan, Schwartz, Joshi, Singh, and
  Smith]{peters-etal-2019-knowledge}
Matthew~E. Peters, Mark Neumann, Robert Logan, Roy Schwartz, Vidur Joshi,
  Sameer Singh, and Noah~A. Smith.
\newblock Knowledge enhanced contextual word representations.
\newblock In \emph{Proceedings of the 2019 Conference on Empirical Methods in
  Natural Language Processing and the 9th International Joint Conference on
  Natural Language Processing (EMNLP-IJCNLP)}, pp.\  43--54, Hong Kong, China,
  November 2019. Association for Computational Linguistics.
\newblock \doi{10.18653/v1/D19-1005}.
\newblock URL \url{https://www.aclweb.org/anthology/D19-1005}.

\bibitem[Radev et~al.(2020)Radev, Zhang, Rau, Sivaprasad, Hsieh, Rajani, Tang,
  Vyas, Verma, Krishna, Liu, Irwanto, Pan, Rahman, Zaidi, Mutuma, Tarabar,
  Gupta, Yu, Tan, Lin, Xiong, and Socher]{radev2020dart}
Dragomir Radev, Rui Zhang, Amrit Rau, Abhinand Sivaprasad, Chiachun Hsieh,
  Nazneen~Fatema Rajani, Xiangru Tang, Aadit Vyas, Neha Verma, Pranav Krishna,
  Yangxiaokang Liu, Nadia Irwanto, Jessica Pan, Faiaz Rahman, Ahmad Zaidi,
  Murori Mutuma, Yasin Tarabar, Ankit Gupta, Tao Yu, Yi~Chern Tan, Xi~Victoria
  Lin, Caiming Xiong, and Richard Socher.
\newblock Dart: Open-domain structured data record to text generation.
\newblock \emph{arXiv preprint arXiv:2007.02871}, 2020.

\bibitem[Shi et~al.(2018)Shi, Tatwawadi, Chakrabarti, Mao, Polozov, and
  Chen]{shi2018incsql}
Tianze Shi, Kedar Tatwawadi, Kaushik Chakrabarti, Yi~Mao, Oleksandr Polozov,
  and Weizhu Chen.
\newblock Incsql: Training incremental text-to-sql parsers with
  non-deterministic oracles.
\newblock \emph{arXiv preprint arXiv:1809.05054}, 2018.

\bibitem[Wang et~al.(2019)Wang, Titov, and Lapata]{wang2019}
Bailin Wang, Ivan Titov, and Mirella Lapata.
\newblock Learning semantic parsers from denotations with latent structured
  alignments and abstract programs.
\newblock In \emph{Proceedings of EMNLP}, 2019.

\bibitem[Wang et~al.(2020)Wang, Shin, Liu, Polozov, and
  Richardson]{wang-etal-2020-rat}
Bailin Wang, Richard Shin, Xiaodong Liu, Oleksandr Polozov, and Matthew
  Richardson.
\newblock {RAT-SQL}: Relation-aware schema encoding and linking for
  text-to-{SQL} parsers.
\newblock In \emph{Proceedings of the 58th Annual Meeting of the Association
  for Computational Linguistics}, pp.\  7567--7578, Online, July 2020.
  Association for Computational Linguistics.
\newblock \doi{10.18653/v1/2020.acl-main.677}.
\newblock URL \url{https://www.aclweb.org/anthology/2020.acl-main.677}.

\bibitem[Wang et~al.(2015{\natexlab{a}})Wang, Berant, and
  Liang]{wang2015parser}
Yushi Wang, Jonathan Berant, and Percy Liang.
\newblock Building a semantic parser overnight.
\newblock In \emph{Proceedings of the 53rd Annual Meeting of the Association
  for Computational Linguistics and the 7th International Joint Conference on
  Natural Language Processing (Volume 1: Long Papers)}, pp.\  1332--1342,
  Beijing, China, July 2015{\natexlab{a}}. Association for Computational
  Linguistics.
\newblock \doi{10.3115/v1/P15-1129}.
\newblock URL \url{https://www.aclweb.org/anthology/P15-1129}.

\bibitem[Wang et~al.(2015{\natexlab{b}})Wang, Berant, Liang,
  et~al.]{wang2015building}
Yushi Wang, Jonathan Berant, Percy Liang, et~al.
\newblock Building a semantic parser overnight.
\newblock In \emph{ACL (1)}, pp.\  1332--1342, 2015{\natexlab{b}}.

\bibitem[Wolf et~al.(2019)Wolf, Debut, Sanh, Chaumond, Delangue, Moi, Cistac,
  Rault, Louf, Funtowicz, and Brew]{Wolf2019HuggingFacesTS}
Thomas Wolf, Lysandre Debut, Victor Sanh, Julien Chaumond, Clement Delangue,
  Anthony Moi, Pierric Cistac, Tim Rault, R'emi Louf, Morgan Funtowicz, and
  Jamie Brew.
\newblock Huggingface's transformers: State-of-the-art natural language
  processing.
\newblock \emph{ArXiv}, abs/1910.03771, 2019.

\bibitem[Yin et~al.(2020{\natexlab{a}})Yin, Neubig, Yih, and
  Riedel]{yin-etal-2020-tabert}
Pengcheng Yin, Graham Neubig, Wen-tau Yih, and Sebastian Riedel.
\newblock {T}a{BERT}: Pretraining for joint understanding of textual and
  tabular data.
\newblock In \emph{Proceedings of the 58th Annual Meeting of the Association
  for Computational Linguistics}, pp.\  8413--8426, Online, July
  2020{\natexlab{a}}. Association for Computational Linguistics.
\newblock \doi{10.18653/v1/2020.acl-main.745}.

\bibitem[Yin et~al.(2020{\natexlab{b}})Yin, Neubig, Yih, and
  Riedel]{yin2020tabert}
Pengcheng Yin, Graham Neubig, Wen-tau Yih, and Sebastian Riedel.
\newblock Tabert: Pretraining for joint understanding of textual and tabular
  data.
\newblock \emph{arXiv preprint arXiv:2005.08314}, 2020{\natexlab{b}}.

\bibitem[Yu et~al.(2018{\natexlab{a}})Yu, Yasunaga, Yang, Zhang, Wang, Li, and
  Radev]{Yu18syntax}
Tao Yu, Michihiro Yasunaga, Kai Yang, Rui Zhang, Dongxu Wang, Zifan Li, and
  Dragomir Radev.
\newblock Syntaxsqlnet: Syntax tree networks for complex and cross-domain
  text-to-sql task.
\newblock In \emph{Proceedings of EMNLP}. Association for Computational
  Linguistics, 2018{\natexlab{a}}.

\bibitem[Yu et~al.(2018{\natexlab{b}})Yu, Zhang, Yang, Yasunaga, Wang, Li, Ma,
  Li, Yao, Roman, Zhang, and Radev]{Yu18emnlp}
Tao Yu, Rui Zhang, Kai Yang, Michihiro Yasunaga, Dongxu Wang, Zifan Li, James
  Ma, Irene Li, Qingning Yao, Shanelle Roman, Zilin Zhang, and Dragomir Radev.
\newblock Spider: A large-scale human-labeled dataset for complex and
  cross-domain semantic parsing and text-to-sql task.
\newblock In \emph{EMNLP}, 2018{\natexlab{b}}.

\bibitem[Zettlemoyer \& Collins(2005)Zettlemoyer and Collins]{Zettlemoyer05}
Luke~S. Zettlemoyer and Michael Collins.
\newblock Learning to map sentences to logical form: Structured classification
  with probabilistic categorial grammars.
\newblock \emph{UAI}, 2005.

\bibitem[Zhang et~al.(2019{\natexlab{a}})Zhang, Zhang, and Balog]{Zhang2019tab}
Li~Zhang, Shuo Zhang, and Krisztian Balog.
\newblock Table2vec: Neural word and entity embeddings for table population and
  retrieval.
\newblock In \emph{Proceedings of the 42Nd International ACM SIGIR Conference
  on Research and Development in Information Retrieval}, SIGIR'19, pp.\
  1029--1032, New York, NY, USA, 2019{\natexlab{a}}. ACM.

\bibitem[Zhang et~al.(2019{\natexlab{b}})Zhang, Yu, Er, Shim, Xue, Lin, Shi,
  Xiong, Socher, and Radev]{zhang19}
Rui Zhang, Tao Yu, He~Yang Er, Sungrok Shim, Eric Xue, Xi~Victoria Lin, Tianze
  Shi, Caiming Xiong, Richard Socher, and Dragomir Radev.
\newblock Editing-based sql query generation for cross-domain context-dependent
  questions.
\newblock In \emph{Proceedings of the 2019 Conference on Empirical Methods in
  Natural Language Processing and 9th International Joint Conference on Natural
  Language Processing}. Association for Computational Linguistics,
  2019{\natexlab{b}}.

\bibitem[Zhong et~al.(2017)Zhong, Xiong, and Socher]{Zhong2017}
Victor Zhong, Caiming Xiong, and Richard Socher.
\newblock Seq2sql: Generating structured queries from natural language using
  reinforcement learning.
\newblock \emph{CoRR}, abs/1709.00103, 2017.

\bibitem[Zhong et~al.(2020)Zhong, Lewis, Wang, and
  Zettlemoyer]{Zhong2020GroundedAF}
Victor Zhong, M.~Lewis, Sida~I. Wang, and Luke Zettlemoyer.
\newblock Grounded adaptation for zero-shot executable semantic parsing.
\newblock \emph{The 2020 Conference on Empirical Methods in Natural Language
  Processing}, 2020.

\end{thebibliography}
\bibliographystyle{iclr2021_conference}

\appendix
\section{Appendices}
\label{sec:appendix}

\begin{table*}[hbt!]
\centering
\scalebox{0.76}{
\begin{tabular}{p{37mm}p{58mm}p{40mm}p{57mm}}
\Xhline{2\arrayrulewidth}
Task & Question & Table/Database & Annotation  \\ \hline
\textsc{Spider} & Find the first and last names of the students who are living in the dorms that have a TV Lounge as an amenity. & database with 5 tables \newline e.g.{\small\tt student}, {\small\tt dorm\_amenity}, ... & \tiny \textsc{SELECT T1.fname, T1.lname FROM student AS T1 JOIN lives\_in AS T2 ON T1.stuid=T2.stuid WHERE T2.dormid IN ( SELECT T3.dormid FROM has\_amenity AS T3 JOIN dorm\_amenity AS T4 ON T3.amenid=T4.amenid WHERE T4.amenity\_name= 'TV Lounge')} \\ \hline
Fully-sup. \textsc{WikiSQL} & How many CFL teams are from York College? & a table with 5 columns \newline e.g. {\small\tt player}, {\small\tt position}, ... & \tiny \textsc{SELECT COUNT CFL Team FROM CFLDraft WHERE College = 'York'}\\ \hline
\textsc{WikiTableQuestions} & In what city did Piotr's last 1st place finish occur? & a table with 6 columns \newline e.g. {\small\tt year}, {\small\tt event}, ... & ``Bangkok, Thailand" \\ \hline
Weakly-sup. \textsc{WikiSQL} & How many CFL teams are from York College? & a table with 5 columns \newline e.g. {\small\tt player}, {\small\tt position},... & 2 \\
\bottomrule
\end{tabular}}
\caption{
Examples of the inputs and annotations for four semantic parsing tasks. 
\textsc{Spider} and Fully-sup. \textsc{WikiSQL} require full annotation of SQL programs, whereas \textsc{WikiTableQuestions} and Weakly-sup. \textsc{WikiSQL} only requires annotation of answers (or denotations) of questions. 
} 
\label{tb:examples}
\end{table*}
\begin{table*}[hbt!]
\centering
\scalebox{0.8}{
\begin{tabular}{rccc}
\toprule
 & Train Size & \# Table & Task \\ \hline
TabFact & 92.2K & 16K & Table-based fact verification \\ \hline
LogicNLG & 28.5K & 7.3K & Table-to-text generation \\ \hline
HybridQA & 63.2K & 13K & Multi-hop question answering \\ \hline
WikiSQL & 61.3K & 24K & Text-to-SQL generation \\ \hline
WikiTableQuestions & 17.6K & 2.1K & Question answering \\ \hline
ToTTo & 120K & 83K & Table-to-text generation \\ \hline
Spider & 8.7K & 1K & Text-to-SQL generation \\ 
\bottomrule
\end{tabular}
}
\caption{Aggregated datasets for table-and-language tasks.} 
\label{tb:agg_data}
\end{table*}

\subsection{Additional Analysis}
\label{sec:add_analysis}

\begin{figure}[hbt!]
    \vspace{-1.5mm}\hspace{-1mm}
    \centering
    \includegraphics[width=0.48\textwidth]{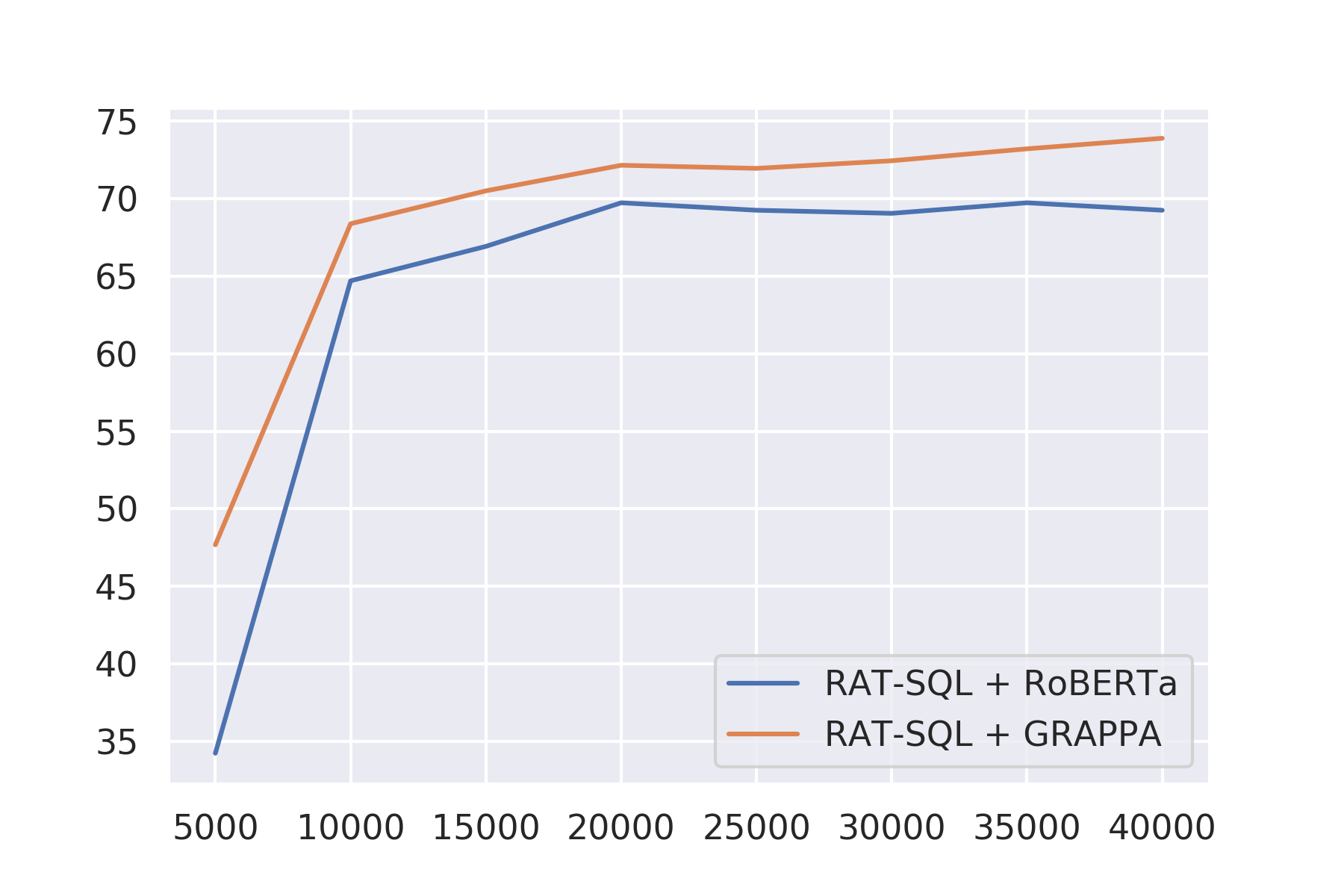}
    \caption{The development exact set match score in \spider{} vs. the number of training steps. RAT-SQL initialized with our pre-trained \tablebert{} converges to higher scores in a shorter time than RAT-SQL $w.$ BERT.}
\label{fig:traincurve}
\vspace{-1mm}
\end{figure}

\paragraph{Training coverage}
As shown in Figure \ref{fig:traincurve}, on the challenging end text-to-SQL \spider{} task, RAT-SQL initialized with \tablebert{} outperforms RAT-SQL using RoBERTa by about 14\% in the early training stage.
This shows that \tablebert{} already captures some semantic knowledge in pre-training.
Finally, \tablebert{} is able to keep the competitive edge by 4\%.

\paragraph{What if the task-specific training data is also used with the MLM or SSP objective in pre-training?}
Although we did not do the same experiments, we would like to point to the RAT-SQL paper~\citep{wang-etal-2020-rat} for some suggestions. They add a similar alignment loss (similar to SSP) on the \spider{} training data and found that it doesn't make a statistically significant difference (in Appendix B).

\begin{figure*}[t]
    \centering
    \subfloat[RoBERTa-large]{
        \includegraphics[width=0.95\linewidth]{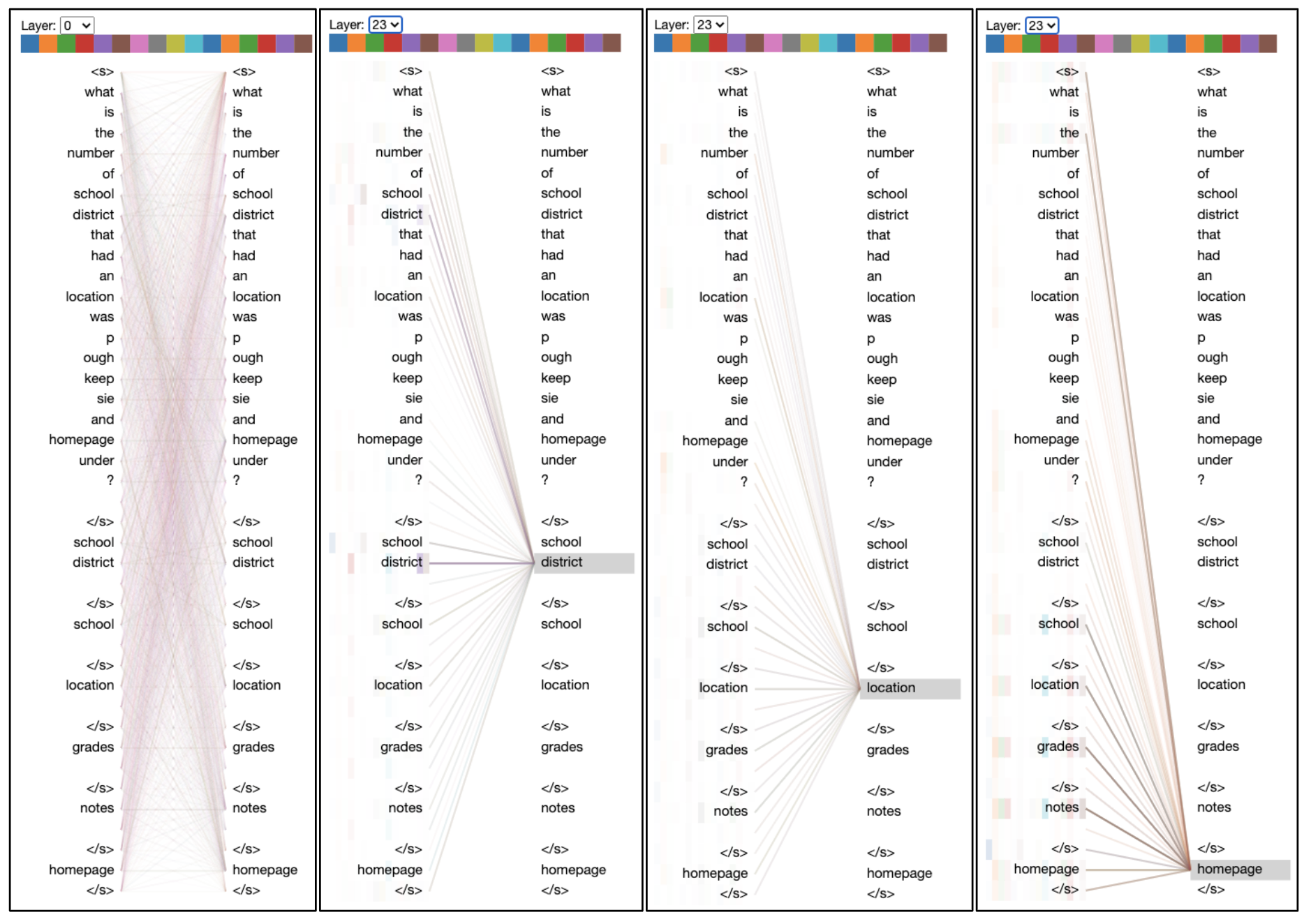}}
    \hfill
    \subfloat[GRAPPA]{
        \includegraphics[width=0.95\linewidth]{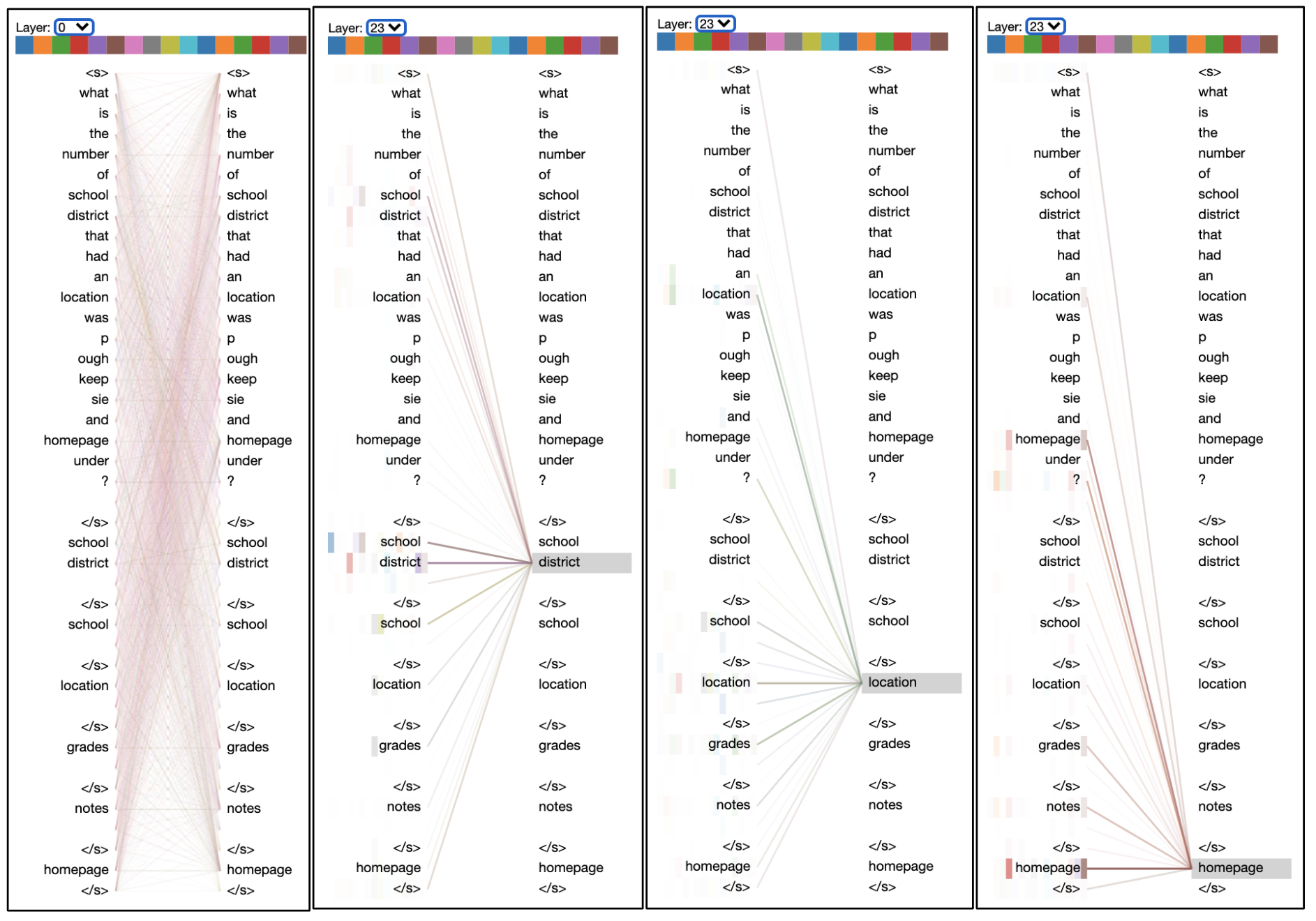}}

    \caption{Attention visualization on the last self-attention layer.}
    \label{fig:attn_viz2}
\end{figure*}

\begin{figure*}[t]
    \centering
    \subfloat[RoBERTa-large]{
        \includegraphics[width=1.0\linewidth]{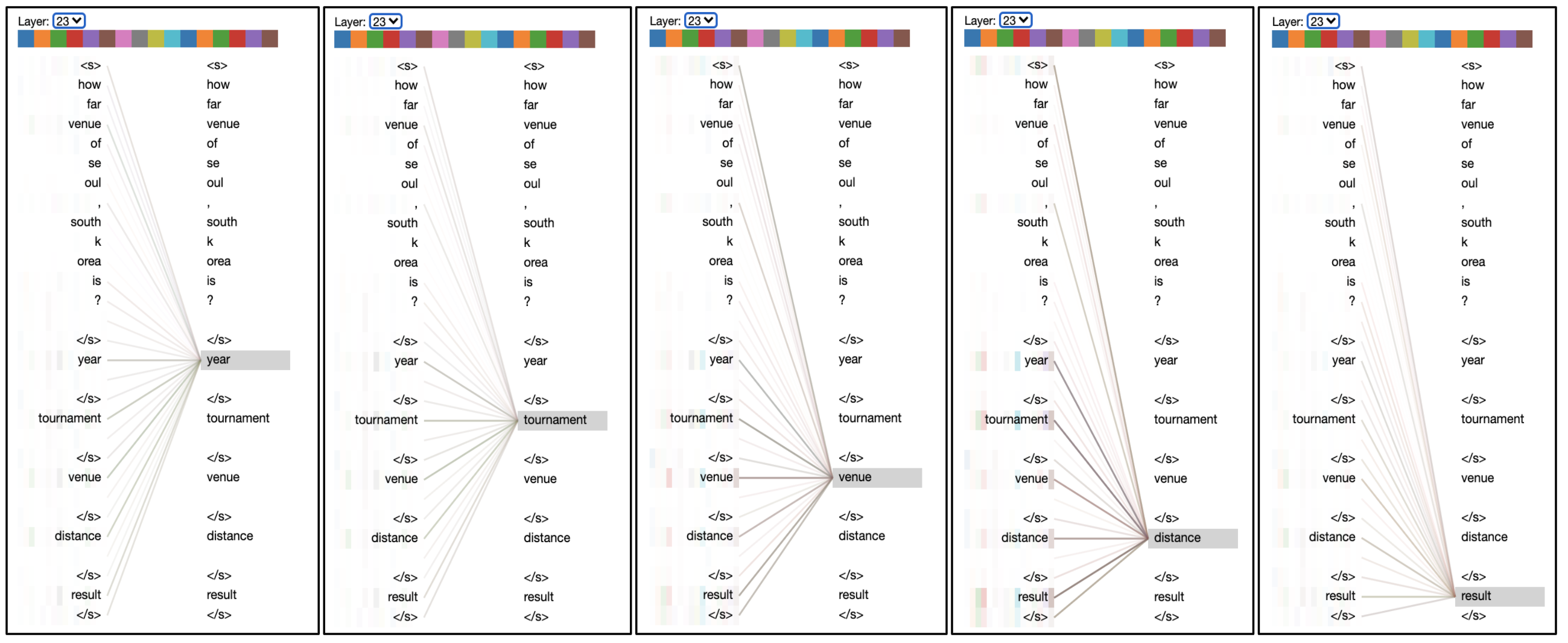}}
    \hfill
    \subfloat[GRAPPA]{
        \includegraphics[width=1.0\linewidth]{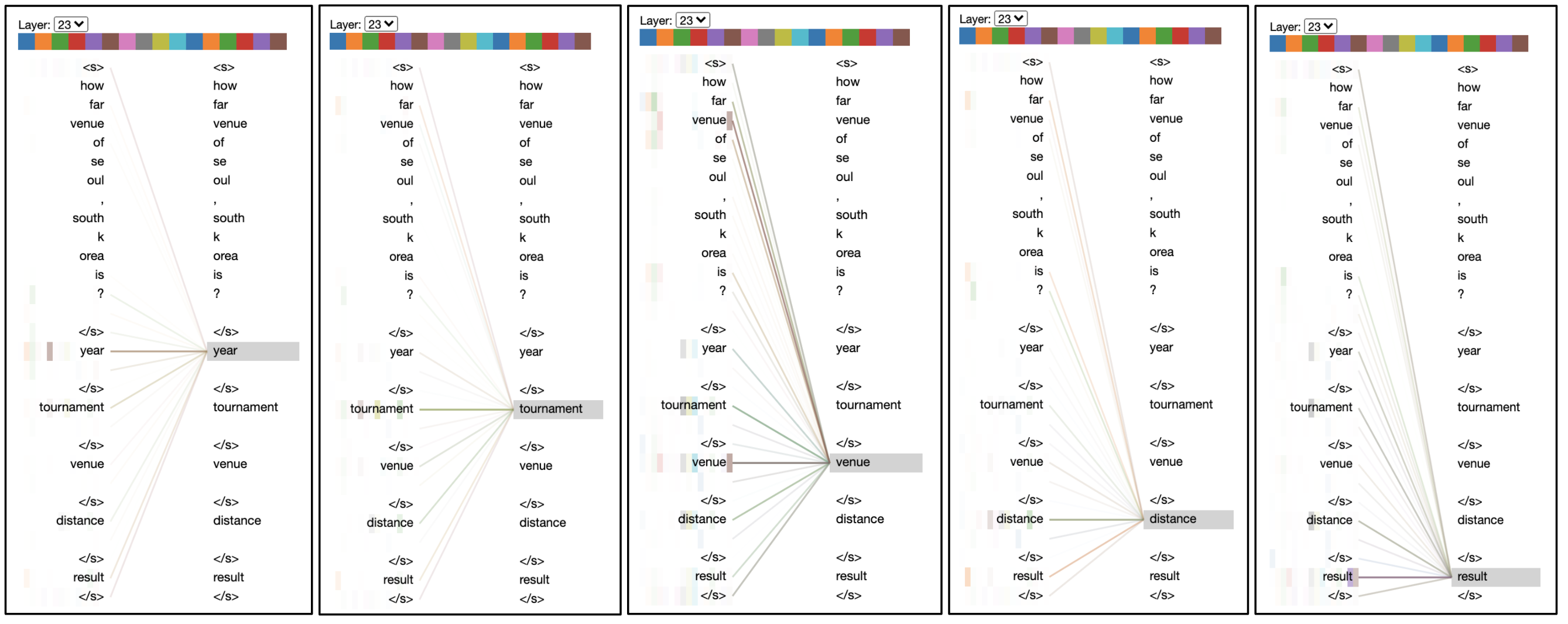}}

    \caption{Attention visualization on the last self-attention layer.}
    \label{fig:attn_viz1}
\end{figure*}

\end{document}